\documentclass[lettersize,journal]{IEEEtran}
\usepackage{amsmath,amsfonts}
\usepackage{algorithmic}
\usepackage{algorithm}
\usepackage{array}
\usepackage[caption=false,font=normalsize,labelfont=sf,textfont=sf]{subfig}
\usepackage{textcomp}
\usepackage{stfloats}
\usepackage{url}
\usepackage{verbatim}
\usepackage{graphicx}
\usepackage{cite}
\usepackage{booktabs}
\usepackage{multirow}
\usepackage[table]{xcolor}
\begin{document}

\title{MSGFusion: Multimodal Scene Graph-Guided Infrared and Visible Image Fusion}

\author{
    \IEEEauthorblockN{Guihui Li$^{1,\dagger}$, Bowei Dong$^{1,\dagger}$,  Kaizhi Dong$^{2}$, Jiayi Li$^{1}$, Haiyong Zheng$^{1,*}$} \\
    \IEEEauthorblockA{$^{1}$College of Computer Science and Technology, Ocean University of China} \\
    \IEEEauthorblockA{$^{2}$College of Electronic Engineering, Ocean University of China} \\
     guihuilee@stu.ouc.edu.cn, dbw@stu.ouc.edu.cn, dongkaizhi@stu.ouc.edu.cn, jiayilee@stu.ouc.edu.cn, zhenghaiyong@ouc.edu.cn
    \thanks{$^{\dagger}$These authors contributed equally to this work, $^{*}$Corresponding authors}

}


\markboth{Journal of \LaTeX\ Class Files,~Vol.~14, No.~8, August~2021}%
{Shell \MakeLowercase{\textit{et al.}}: A Sample Article Using IEEEtran.cls for IEEE Journals}

\maketitle

\begin{abstract}
Infrared and visible image fusion has garnered considerable attention owing to the strong complementarity of these two modalities in complex, harsh environments. While deep learning–based fusion methods have made remarkable advances in feature extraction, alignment, fusion, and reconstruction, they still depend largely on low-level visual cues, such as texture and contrast, and struggle to capture the high-level semantic information embedded in images. Recent attempts to incorporate text as a source of semantic guidance have relied on unstructured descriptions that neither explicitly model entities, attributes, and relationships nor provide spatial localization, thereby limiting fine-grained fusion performance. To overcome these challenges, we introduce MSGFusion, a multimodal scene graph–guided fusion framework for infrared and visible imagery. By deeply coupling structured scene graphs derived from text and vision, MSGFusion explicitly represents entities, attributes, and spatial relations, and then synchronously refines high-level semantics and low-level details through successive modules for scene graph representation, hierarchical aggregation, and graph-driven fusion. Extensive experiments on multiple public benchmarks show that MSGFusion significantly outperforms state-of-the-art approaches, particularly in detail preservation and structural clarity, and delivers superior semantic consistency and generalizability in downstream tasks such as low-light object detection, semantic segmentation, and medical image fusion.
\end{abstract}

\begin{IEEEkeywords}
Image fusion, infrared and visible images, vision and language, visual Scene Graph, textual Scene Graph
\end{IEEEkeywords}



\section{Introduction}
\begin{figure}[tb]
	{\centering\includegraphics[width=0.5\textwidth]{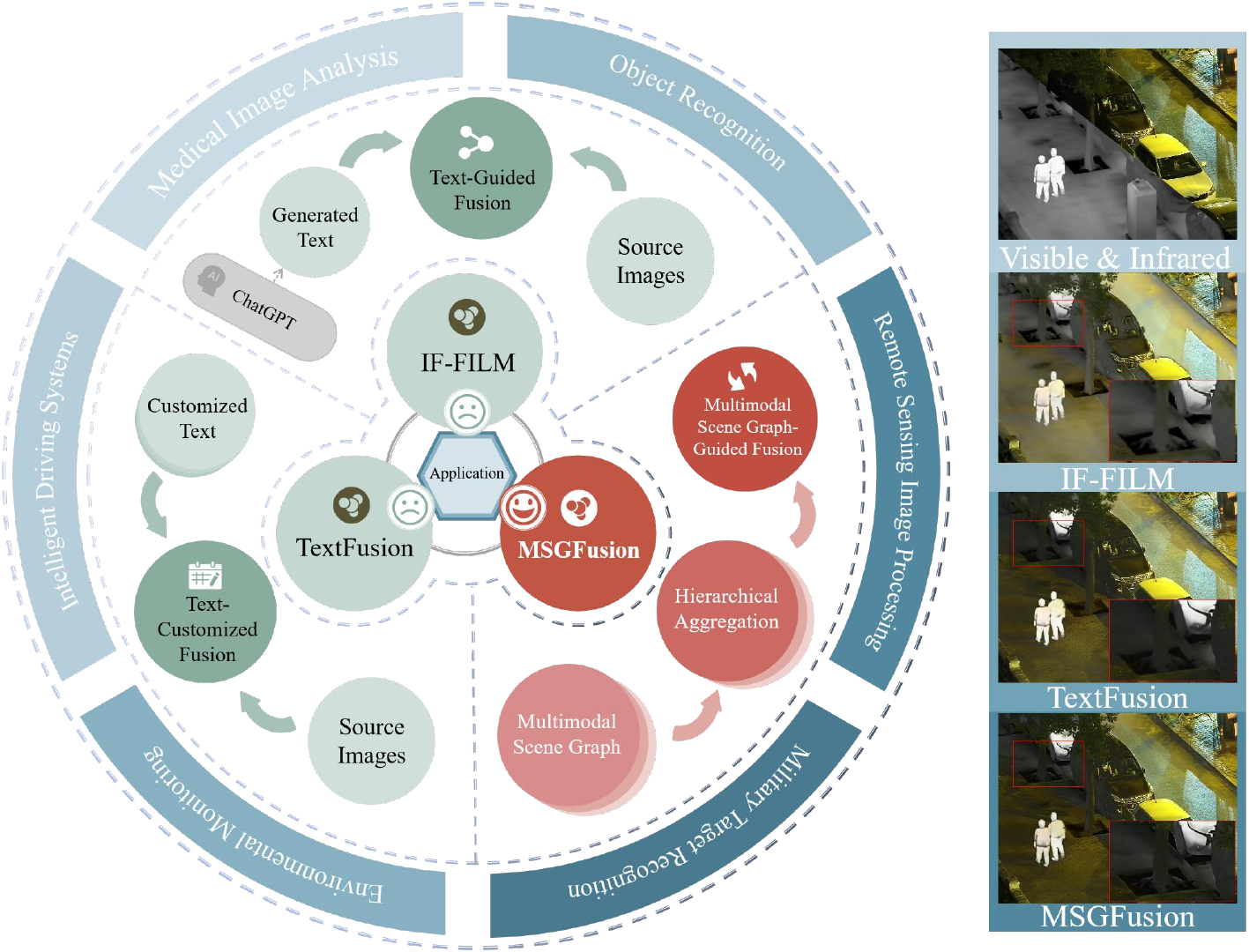}}
    \caption{Illustrations of our core idea. As shown, most existing methods rely on unstructured text prompts for semantic guidance, which necessitates generating or manually refining textual descriptions and fails to explicitly model entities, attributes, and spatial relationships. Differently, our approach adopts a deeply structured coupling of textual and visual scene graphs: it first constructs modality‐specific scene graph representations, then performs hierarchical aggregation, and finally employs a graph‐driven fusion module to synchronously optimize high‐level semantics and low‐level details, achieving precise cross‐modal alignment and high‐quality fusion.}
	\label{fig1-1}
\end{figure}

Image fusion integrates data from multiple sensors capturing the same scene to produce a more comprehensive and reliable representation. In intelligent perception systems, infrared and visible images serve as two key modalities. As illustrated in Figure \ref{fig1-1}, they hold significant application value in object recognition~\cite{wang2023interactively,liu2022target}, environmental monitoring~\cite{slonecker2010visible}, and security surveillance~\cite{krotosky2008person}.

\begin{figure*}[h]
  \centering
  \includegraphics[width=1\textwidth]{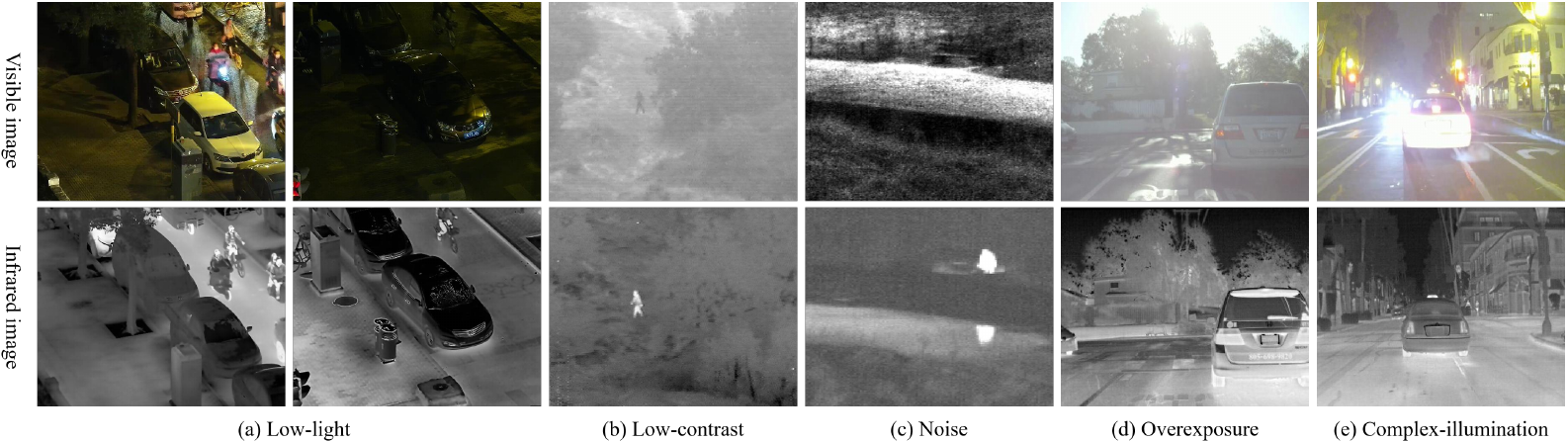}
  \caption{(A) A typical example of domain entanglement between multi-exposure and multi-focus domains and qualitative comparison among different methods. (B) Feature space visualization of a typical unified fusion method and our method.}
  \label{fig1}
\end{figure*}

Spectrally, the infrared band (760 nm to 1 mm) lies just beyond the red end of the visible spectrum; its radiometric characteristics are governed by an object’s thermodynamic state, as materials absorb and re-emit infrared radiation, allowing infrared imaging to accurately reflect temperature distributions and thermal properties. By contrast, the visible band (380 nm to 760 nm) depends on ambient illumination to deliver high-resolution spatial detail and rich texture information, yet remains highly susceptible to lighting conditions. Although single-band imagery inherently struggles in complex environments, infrared and visible images exhibit strong complementarity in environmental adaptability and visual features. As illustrated in Fig.~\ref{fig1}, visible images exhibit abundant texture details but are vulnerable to complex illumination changes, adverse weather, and occlusions, whereas infrared images supply stable thermal radiation information across all lighting conditions yet suffer from insufficient texture detail and may introduce thermal noise and contrast loss. Through multispectral fusion, these modalities can be synergistically combined to enhance both robustness and information integrity, thereby providing more reliable data support for applications such as remote sensing~\cite{price1997spectral}, military reconnaissance~\cite{gerken2017military}, and autonomous driving~\cite{li2023emergent}.

Traditional image fusion techniques originally relied on mathematical transforms and hand-crafted fusion rules to produce fused images~\cite{ma2019infrared,zhang2023visible}. However, manually designed features often introduce redundancy, and fixed rule-based schemes struggle to adapt to complex, dynamic scenes. In recent years, deep learning–based approaches have been widely adopted in the image fusion field, consistently demonstrating superior visual quality, robustness, and computational efficiency compared to classical methods, and have thus attracted significant attention.

Although deep learning–based methods have made remarkable strides in feature extraction~\cite{zhang2017infrared}, alignment~\cite{li2023feature}, fusion~\cite{tang2022piafusion}, and reconstruction~\cite{su2022infrared}, the vast majority still depend primarily on low-level visual cues, such as texture and contrast, while paying insufficient attention to the deep, high-level semantic information inherent in the images. Some efforts have attempted to integrate downstream tasks, such as semantic segmentation~\cite{tang2022image} or object detection~\cite{liu2022target}, into the fusion pipeline. Yet, they remain confined to pixel-level semantics and fail to fully harness the potential of semantic guidance.

Recently, large-scale vision–language pretraining models such as CLIP~\cite{tu2023closer} and GPT-4~\cite{achiam2023gpt} have demonstrated extraordinary capabilities in cross-modal semantic understanding and generation. As illustrated in Figure~\ref{fig1-1}, one line of research leverages ChatGPT to generate text descriptions that steer the fusion process~\cite{IF-FILM}; another employs manually crafted prompts in conjunction with CLIP to establish a coarse-to-fine semantic alignment mechanism for controllable image fusion~\cite{TextFusion,Text-IF}; yet others define fusion objectives in IVIF via textual descriptions and encode them into a multimodal embedding space using CLIP~\cite{wang2024infrared}.

However, directly integrating text into image fusion presents several critical challenges. First, unstructured textual descriptions lack explicit modeling of entities, attributes, and relationships; they merely narrate scene elements in natural language, leading to redundant information and difficulty in emphasizing key components. Second, texts generated by different large language models exhibit substantial variability in style, level of detail, and focal points, making it hard to ensure both comprehensiveness and consistency. Third, textual data inherently lacks spatial localization and the fine-grained texture structure of image objects, complicating direct alignment with pixel-level features. Consequently, discovering and leveraging deep, fine-grained semantic features that go beyond purely visual and textual representations remains a central research challenge.

Scene graphs, as a structured semantic representation, explicitly encode entities, attributes and their interrelations, thereby capturing both local and global scene context. Compared to global text embeddings, textual scene graphs furnish a more logically coherent semantic prior, while visual scene graphs quantitatively model object-to-object spatial relationships and low-level visual attributes, enabling precise extraction and fusion of complementary multimodal information. To overcome the challenges outlined above, we propose MSGFusion, a novel multimodal scene graph–guided image fusion framework, as illustrated in Figure.~\ref{fig1-1}, for the first time, deeply integrates structured semantics from both visual and textual modalities. Our multimodal scene graph module simultaneously extracts high-level conceptual semantics from text and authentic visual cues (including spatial relations and appearance attributes) from infrared and visible images, ensuring that the fusion process preserves both abstract meaning and fine-grained detail.

Specifically, MSGFusion first constructs modality-specific scene graph embeddings via the Multimodal Scene Graph Representation module. These embeddings are then semantically fused through the hierarchical guided aggregation module, yielding a unified scene graph embedding. Finally, the scene graph–driven fusion module leverages this embedding to produce a high-quality fusion image. Extensive experiments on multiple public benchmarks show that MSGFusion consistently outperforms leading fusion methods, especially in terms of detail preservation, semantic consistency, and structural clarity. Furthermore, MSGFusion not only accelerates downstream task performance but also demonstrates exceptional generalizability in medical image fusion applications.

Our contributions can be summarized as follows:
\begin{itemize}
 \item We introduce MSGFusion, the first framework to deeply couple textual conceptual semantics with visual attributes and spatial relationships from infrared and visible images, enabling fusion that simultaneously preserves high-level semantics and low-level details.
 \item We design a multimodal scene graph representation module and a hierarchically guided aggregation module to uniformly abstract entities, attributes, and relations into a fine-grained multimodal semantic graph, providing rich and actionable semantic priors for flexible fusion.
 \item We propose a scene graph–driven fusion module that adaptively modulates visual feature fusion strategies based on the multimodal scene graph, significantly enhancing detail preservation, semantic consistency, and structural clarity in the fused output.
\end{itemize}

\section{Related Work}
\subsection{Deep Learning-based Infrared and Visible Image Fusion}
Currently, deep learning-based infrared and visible image fusion methods fall into four categories: AE~\cite{li2021rfn,li2018densefuse}, CNN~\cite{zhang2021sdnet, zhang2020rethinking}, GAN~\cite{ma2019fusiongan,ma2020ddcgan} and Transformer~\cite{ma2022swinfusion, park2023cross}. Specifically, AE‐based approaches leverage an encoder for feature extraction and reconstruction, apply handcrafted or modular fusion strategies, and employ a decoder to produce the fused output. For instance, Li et al.\cite{li2018densefuse} propose dense‐block encoder–decoder architecture that first fuses deep features from two source images via a handcrafted fusion rule, and then refines the result with adaptive fusion using a residual fusion network\cite{li2021rfn}. Zhao et al.\cite{zhao2021didfuse} propose a dual‐branch autoencoder that decomposes input into modality‐specific and shared features, later extending this framework to a more powerful variant\cite{zhao2023cddfuse}. More recently, Luo et al.~\cite{LUO2024110192} develop a hierarchical encoder–decoder network augmented with cascaded edge priors to address low‐contrast and blurred edge details. Despite their impressive results, these AE‐based fusion methods still rely on manually defined fusion strategies and, due to their dependence on pretrained models, may suffer from suboptimal feature representations. 

Consequently, several works explore end‐to‐end image fusion networks built on convolutional neural networks. For example, Zhang et al.\cite{zhang2020rethinking} develop a dual‐branch architecture for unified fusion and further enhance details via a compression–decompression network\cite{zhang2021sdnet}. Xu et al.\cite{xu2020u2fusion} propose an end‐to‐end fusion framework with an adaptive loss function designed to preserve source image information. Huang et al.\cite{huangreconet} introduce a recursive refinement network for efficient multimodal fusion, while Liu et al.~\cite{LIU2024110226} design an adaptive feature selection module to filter feature maps and boost fusion performance. These CNN‐based methods typically rely on sophisticated network architectures and tailor‐made loss functions to enable effective end‐to‐end training.

Additionally, several studies frame image fusion as a minimax game between a generator and a discriminator, constraining the probability distributions of the fused and source images to enforce rich texture detail. GAN-based methods thus cast fusion as an adversarial process: Ma et al.\cite{ma2019fusiongan} introduce a GAN architecture for infrared and visible fusion but encounter imbalance due to a single discriminator, an issue subsequently resolved by DDcGAN\cite{ma2020ddcgan}. Liu et al.\cite{liu2022target} propose a bilevel optimization framework that incorporates a detection loss, while Rao et al.\cite{rao2023gan} develop a GAN augmented with intensity attention and semantic transition modules to extract key information.

Transformer-based fusion methods have attracted widespread attention due to their ability to capture long-range dependencies and global interactions. Ma et al.\cite{ma2022swinfusion} design a Swin Transformer–based network for cross-domain learning, while Park et al.\cite{park2023cross} introduce a cross-modal fusion algorithm tailored for infrared and visible fusion. Chang et al.\cite{chang2023aft} employ a Transformer to model multimodal relationships and interactions, and Yang et al.\cite{YANG2024110223} develop SePT, a semantic-aware Transformer that better preserves texture in fused outputs. Contemporary Transformer-based approaches often integrate CNN backbones, leveraging self-attention or multi-scale schemes to fuse features. These methods are predominantly unsupervised, optimizing a fusion loss between the fused image and its sources, with a strong emphasis on global context modeling.

Although these methods have achieved notable gains in feature extraction, alignment, fusion, and reconstruction, they still depend heavily on low-level visual cues, such as texture and contrast, while overlooking the rich, high-level semantic information embedded in images. Moreover, many approaches require training separate models for different downstream tasks, which limits both the flexibility and controllability of the fusion output~\cite{luo2024infrared}. This absence of adaptive control mechanisms constrains their applicability across diverse, real-world scenarios.

To address this shortcoming, recent research has begun to integrate text semantics into the fusion process. For instance, Wang et al.~\cite{wang2024infrared} formulate IVIF objectives in natural language and use CLIP to encode the corresponding text into a unified multimodal embedding space. Cheng et al.~\cite{TextFusion} introduce a coarse-to-fine semantic alignment mechanism based on vision–language pretraining, enabling fully controllable, text-guided image fusion. Zhao et al.~\cite{IF-FILM} propose a novel text–image collaborative fusion model that leverages descriptions generated by large language models to guide the fusion procedure, achieving superior performance across multiple tasks. Yi et al.~\cite{Text-IF} develop an interactive fusion framework combining text-driven semantic guidance with degradation-aware processing, wherein a text encoder and a semantic interaction decoder jointly facilitate dynamic multimodal fusion and user-specified output control.

\subsection{Multimodal Scene Graph}   
\textbf{Visual Scene Graph Representation.}
A Visual Scene Graph (VSG) explicitly represents the objects, attributes and semantic relations in an image through a node–edge structure.  
Since Johnson et al.\,\cite{sg1} first introduce VSGs for image retrieval and captioning, they have become a core foundation for image understanding, visual question answering and cross-modal retrieval.  
Early pipelines typically decouple object detection from relation classification, which limited the exploitation of contextual dependencies.  
Xu et al.\,\cite{sg2} address this limitation with an iterative message-passing network that recursively aggregates node and edge states via graph neural networks, thereby enabling end-to-end joint reasoning of objects and relations.  
Zellers et al.\,\cite{sg3} statistically reveal higher-order motifs in scene graphs and achieved large gains by modelling global context sequentially in the Neural Motifs framework.  
To mitigate bias arising from long-tail distributions, Tang et al.\,\cite{sg4} adopt a causal perspective and used the total direct effect to perform counterfactual interventions on contextual bias, markedly improving recall for rare relations.  
Zhang et al.\,\cite{sg7} further incorporate a hierarchical external knowledge graph (HiKER-SGG), maintaining robust structural reasoning under severe degradations.  
For finer, pixel-level structure, Yang et al.\,\cite{sg8} propose the Panoptic Scene Graph, which jointly models objects, attributes and background atop instance segmentation, realising holistic, panoptic-level scene reasoning.
VSGs also boost cross-modal alignment and generation quality.  
Li et al.\,\cite{sg5} inject scene-graph structure into an image-captioning system via hierarchical attention, improving descriptive accuracy and diversity.  
Chen et al.\,\cite{sg6} introduce Abstract Scene Graphs for fine-grained, controllable caption generation.  
Wu et al.\,\cite{sg9} perform contrastive learning on multi-level scene-graph components within a unified visual–semantic space, yielding stronger fine-grained alignment and adversarial robustness.  
Huang et al.\,\cite{sg10} explicitly infuse scene-graph knowledge into Structure-CLIP and constructed structured negatives, greatly enhancing a large-scale vision–language model’s discrimination of relational semantics.
Although VSG techniques have advanced structured modelling, debiasing, knowledge injection and cross-modal fusion, their potential in image fusion remains unexplored.  
This work therefore introduces explicit VSG semantics into a multimodal image-fusion framework for the first time, aiming to achieve new breakthroughs in detail preservation and semantic consistency.

\textbf{Textual Scene Graph Representation.}
Textual Scene Graph (TSG) is a structured semantic representation that effectively captures entities, attributes, and their intricate semantic relationships present in natural-language descriptions, and has been widely adopted in tasks such as image–text matching, multimodal reasoning, and image generation. Some early studies typically rely on dependency parsers to construct initial scene graph structures. For instance, Schuster et al.\,\cite{text1} introduce a parsing method that integrates dependency-based rules and classifiers to accurately convert sentences into scene graphs, significantly enhancing performance in image retrieval tasks. Johnson et al.\,\cite{text2} further employ graph convolutional networks (GCN) for structured reasoning on textual scene graphs, effectively improving the model's capability to handle complex semantic relationships. To improve semantic accuracy and robustness, Li et al.\,\cite{text3} present the FACTUAL-MR framework, rigorously defining semantic rules for entities, attributes, and relations within scene graphs, thereby substantially enhancing their semantic completeness and consistency. Concurrently, Jiang et al.\,\cite{text4} propose an approach that incorporates hierarchical relational structures and commonsense knowledge validation, effectively mitigating semantic conflicts and commonsense errors during the generation of scene graphs. 
Wu et al.\,\cite{text5} propose a scene graph hallucination diffusion model for synthesizing complex images from abstract textual descriptions, progressively enriching textual scene graph structures through a discrete diffusion process and achieving higher-quality image synthesis. Linok et al.\,\cite{text6} designe a dynamic graph encoder, DyGEnc, which serializes textual scene graphs for dynamic scene reasoning, effectively enhancing the model's performance in dynamic visual question answering tasks. Additionally, Zhao et al.\,\cite{text7} introduce an unsupervised Caption-to-PSG task, leveraging pure image–text pairs to generate pixel-level panoptic scene graphs from text, substantially reducing the dependence of scene graph tasks on densely annotated data. Pham et al.\,\cite{text8} develop the CORA model, which utilizes a two-stage graph attention network to finely aggregate features of objects, attributes, and relations within scene graphs, significantly improving fine-grained semantic alignment performance in image–text retrieval. Chen et al.\,\cite{text9} introduce the SGP framework, which for the first time leverages large language models (LLM) in a role-playing manner to generate scene graphs, significantly enhancing the model's capability in semantic role differentiation and complex situational understanding. 
Inspired by these studies, this work introduces, for the first time, a structured semantic embedding mechanism of textual scene graphs into the image fusion task. By employing a semantic concept encoder and graph attention networks, our approach explicitly models linguistic graph structures, thereby effectively improving the semantic consistency and structural fidelity of the fused images.
\subsection{Comparison with Existing Approaches}
Unlike existing approaches that depend on downstream visual semantics, such as semantic segmentation~\cite{tang2022image} or object detection~\cite{liu2022target}, to guide fusion, MSGFusion is the first to deeply couple structured semantics from infrared, visible, and textual modalities via multimodal scene graphs, achieving fine‐grained alignment and fusion of high‐level concepts with low‐level visual features. Compared to controllable fusion methods that rely solely on global text prompts~\cite{TextFusion,IF-FILM,Text-IF}, our method’s scene graph representation and hierarchical aggregation modules explicitly model entities, attributes, and relationships. This delivers richer local-to-global semantic priors, enables dynamic emphasis on critical regions, and significantly improves detail preservation, semantic consistency, and structural integrity in the fused outputs.

\section{Method}
\begin{figure*}[htb]
	{\centering\includegraphics[width=1\textwidth]{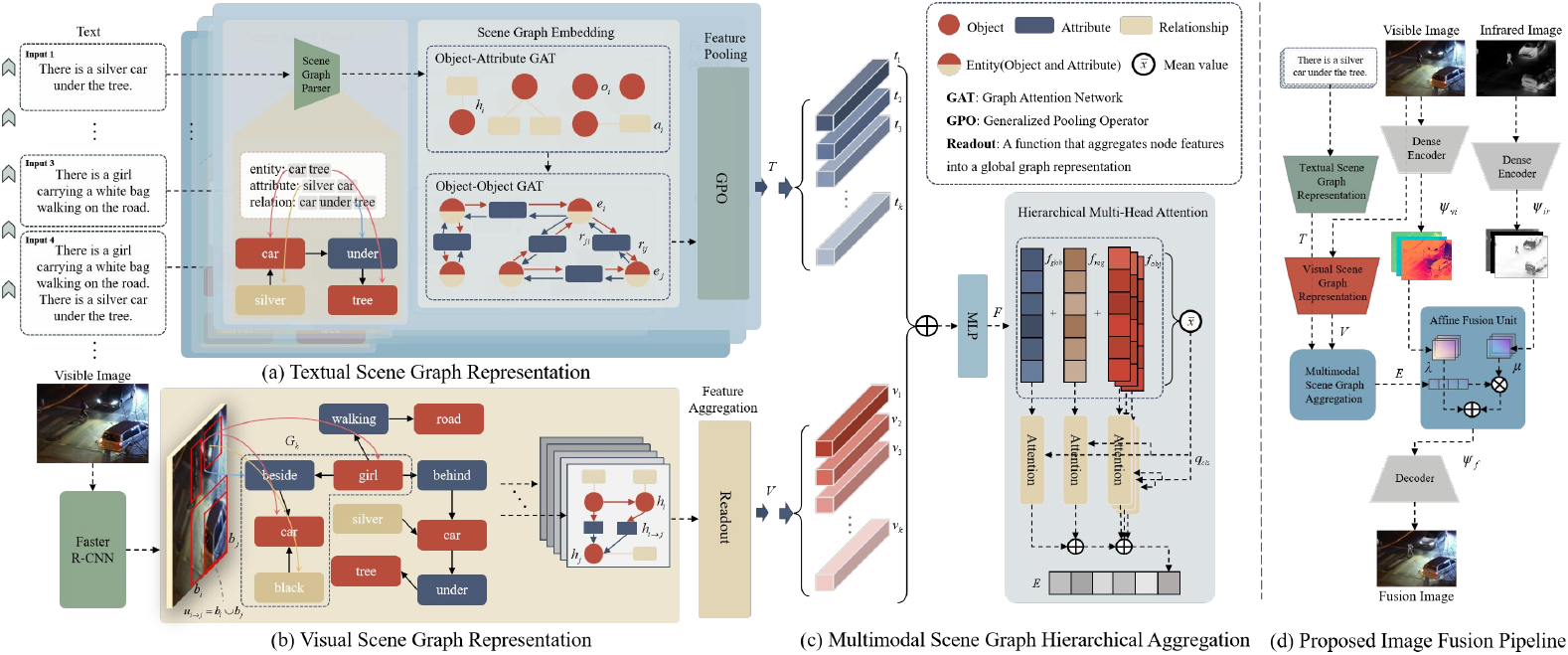}}
    \caption{Overall framework of the proposed multimodal scene graph–driven image fusion model. (a) Textual Scene Graph Representation Module: Takes multi‐level text descriptions as input, uses a scene parser to construct a textual scene graph, and employs a graph embedding network to generate hierarchical text feature embeddings. (b) Visual Scene Graph Representation Module: Processes the visible image through an object detector and a graph neural network to extract candidate regions and produce visual scene graph embeddings. (c) Multimodal Scene Graph Hierarchical Aggregation Module: Aligns and jointly models the textual and visual scene graph embeddings to derive a unified multimodal semantic representation. (d) Scene Graph Driven Fusion Module: Uses the fused semantic embedding to guide the fusion of infrared and visible visual features, producing a high-quality fused image.
    }
	\label{fig3-1}
\end{figure*} 

\subsection{Overall Framework}
Figure~\ref{fig3-1} illustrates the overall architecture of the proposed multimodal scene graph driven image fusion model.  
The model introduces, for the first time, cross-modal scene graphs as a unified semantic representation, structurally modelling and jointly fusing information from the visual and linguistic modalities.  
This design enhances the fused image in terms of semantic consistency, detail preservation, and background integrity.  
The entire system consists of three components: multimodal scene-graph representation, multimodal scene-graph hierarchical aggregation, and a scene-graph-driven fusion module.

Specifically, given an infrared image $I_{ir}$, a visible image $I_{vi}$, and multiple corresponding textual descriptions, the source images are first fed into a Dense Encoder~\cite{li2018densefuse} to extract multi-scale visual features.  
Concurrently, the visible image is processed by Faster~RCNN to obtain candidate object regions; ROI pooling then yields local region features, which are updated via multiple rounds of message passing in a GRU-based graph neural network~\cite{sg2} to construct the visual scene graph $G_{v}$.  
In this graph, nodes denote object entities in the image, and edges encode their spatial and semantic relationships.

For multiple textual descriptions corresponding to the source image, they are transformed into an initial scene graph by scene graph parser~\cite{text1} and semantic embedding of nodes is obtained by semantic concept encoder. 
We further employ an object–attribute graph attention network(GAT) to model entity features composed of object–attribute pairs, followed by an object–object relation GAT to capture inter-entity relations; a generalized pooling operation (GPO) finally yields the textual scene-graph embedding $G_{t}$.  
Subsequently, the multimodal scene-graph hierarchical aggregation module adopts a multimodal hierarchical multi-head attention mechanism to align and fuse the visual and textual modalities across layers, producing a unified multimodal embedding $\mathbf{E}$.
Finally, the scene-graph–driven fusion module treats the multimodal embedding $E$ as the fusion feature: the infrared modality generates the fusion weight term~$\mu$, while the visible modality provides the bias term~$\lambda$, thereby achieving pixel-level fine-grained fusion.  
The overall fusion computation is expressed as: 
\begin{equation}
\begin{split}
\hat{I}_{f} = F(I_{ir},\, I_{vi},\, G_{t},\, G_{v}) = \mu \odot\mathbf{E} + \lambda ,
\end{split}
\end{equation}
where $\odot$ denotes the Hadamard product, $\hat{I}_{f}$ is the final fused image, $\mathbf{E}$ is the high-dimensional fusion feature obtained from the aggregation of visual and textual scene graphs, and $\mu, \lambda$ are the weight and bias terms produced by the infrared and visible modalities, respectively, during the fusion process.
\subsection{Multimodal Scene Graph Representation}
\subsubsection{Visual Scene Graph Representation}
To realize the structured modelling of objects and their semantic relations in an image, the visual modality adopts a scene-graph construction pipeline based on object detection and graph neural reasoning. This module consists of three stages: target area proposals, feature extraction, and scene graph reasoning.

\textbf{Target Area Proposals.}
First, in order to obtain potential object regions in the image, Faster~RCNN is employed as the base object detector, whose core component is a Region Proposal Network (RPN).  
Given a visible image~$I_{vi}$, the RPN predicts, via a multi-scale feature pyramid, a set of candidate bounding boxes
\(
B=\{b_{1}, b_{2}, \ldots, b_{n}\},
\)
where each $b_{i}$ corresponds to a potential object region; these regions subsequently serve as the entity nodes (object–attribute pairs) of the scene-graph structure.

\textbf{Feature Extraction.}
Region-level features are extracted from the backbone feature map for the above candidate boxes to initialise the visual representation of the scene-graph nodes.  
Specifically, for each candidate box~$b_{i}$, ROI pooling is applied to obtain, at a fixed spatial resolution, the corresponding visual feature vector $\mathbf{f}_{i}\!\in\!\mathbb{R}^{d}$ from the backbone feature map~$F$:
\begin{equation}
\begin{split}
\mathbf{f}_{i} = \mathrm{ROI\_pool}(F,\, b_{i}),
\end{split}
\end{equation}
where the vector~$\mathbf{f}_{i}$ captures the semantic representation of the candidate box~$b_{i}$ at its spatial location and is used as the initial node feature fed into the graph neural network.

However, modelling the target nodes alone is not sufficient to capture the structural dependencies and spatial interactions between entities.
To further explore the contextual semantic relations between candidate objects, a union box is constructed for every pair of candidate boxes $(b_{i}, b_{j})$ so as to model their interactive region, defined as: 
\begin{equation}
\begin{split}
u_{i\rightarrow j} = b_{i} \cup b_{j}, 
\end{split}
\end{equation}
and extract features from the joint region based on the visual feature map $F$: 
\begin{equation}
\begin{split}
\mathbf{f}_{i\rightarrow j} = \mathrm{ROI\_pool}(F,u_{i\rightarrow j}). 
\end{split}
\end{equation}

Up to this point, each candidate object pair $(b_{i}, b_{j})$ is endowed with a tri-tuple feature $(\mathbf{f}_{i}, \mathbf{f}_{j}, \mathbf{f}_{i\rightarrow j})$ that represents their semantic entities, interactive target, and spatial relation, thereby providing representational support for the subsequent graph-neural semantic reasoning.

\textbf{Scene Graph Reasoning.}
To realise contextual interaction between object nodes and relation edges, this work introduces an iterative graph neural network (GNN) based on the gated recurrent unit (GRU) as the scene-graph reasoning module.  
In the constructed graph, nodes represent object entities, whereas edges encode semantic relations between entities (e.g., spatial or functional dependencies).  
The module conducts multi-round message passing to update graph states, thereby achieving semantic enhancement and structure awareness.

Let the hidden state of an object node~$v_{i}$ be $h_{i}^{(t)}$ and that of an edge $e_{i\!\rightarrow\!j}$ be $h_{i\!\rightarrow\!j}^{(t)}$.  
The update mechanism at iteration~$t$ is formulated as: 
\begin{equation}
\begin{split}
h_{i}^{(t)} \,=\, \operatorname{GRU}_{node}\!\bigl(m_{i}^{(t-1)},\, h_{i}^{(t-1)}\bigr),\\
h_{i\!\rightarrow\!j}^{(t)} \,=\, \operatorname{GRU}_{edge}\!\bigl(m_{i\!\rightarrow\!j}^{(t-1)},\, h_{i\!\rightarrow\!j}^{(t-1)}\bigr),
\end{split}
\end{equation}
where $\mathbf{m}_{i}^{(t-1)}$ and $\mathbf{m}_{i\!\to\! j}^{(t-1)}$ denote the node level and edge level contextual messages gathered at iteration~$t\!-\!1$.  
To enable adaptive message aggregation, we introduce a gated, weighted pooling mechanism with learnable parameters, defined as follows:
\begin{equation}
\begin{split}
m_{i}^{(t-1)}       \;=\; \sum_{j}\sigma\!\bigl(v_{1}^{\top}[h_{i},\,h_{i\!\rightarrow\!j}]\bigr)\,\cdot\,h_{i\!\rightarrow\!j} \\
             \;+\; \sum_{j}\sigma\!\bigl(v_{2}^{\top}[h_{i},\,h_{j\!\rightarrow\!i}]\bigr)\,\cdot\,h_{j\!\rightarrow\!i},
\end{split}
\end{equation}
\begin{equation}
\begin{split}
m_{i\!\rightarrow\!j}^{(t-1)} = \sigma\!\bigl(w_{1}^{\top}[h_{i},\,h_{i\!\rightarrow\!j}]\bigr)\,\cdot\,h_{i} \\
                      \;+\; \sigma\!\bigl(w_{2}^{\top}[h_{i},\,h_{j\!\rightarrow\!i}]\bigr)\,\cdot\,h_{j},
\end{split}
\end{equation}
where $v_{1}, v_{2}, w_{1}, w_{2}\in\mathbb{R}^{2d}$ are learnable parameters; $[\cdot,\cdot]$ denotes vector concatenation; and $\sigma(\cdot)$ is the sigmoid activation function, which realises the gating mechanism and dynamically adjusts the influence weights of different message sources.

After the multi–round iterations, the hidden states of nodes and edges encapsulate contextual semantic information, thereby forming a complete graph–level embedding.  
To obtain the final visual scene-graph representation, the model filters the graph nodes and selects the top-scoring $n$ object nodes $\{o_{1},o_{2},\dots,o_{n}\}$ together with their incident relations, constructing a series of local sub-graphs:
\begin{equation}
\begin{split}
G_{k} = (o_{k},\,\mathcal{R}_{k}), \qquad k = 1,\dots,n ,
\end{split}
\end{equation}
where $\mathcal{R}_{k}$ denotes the set of semantic edges connecting object $o_{k}$ to the other nodes.  
For each sub-graph $G_{k}$, a Readout Function is introduced to aggregate the hidden states of its nodes and edges, yielding the final embedding vector $\mathbf{v}_{k}\in\mathbb{R}^{d}$:
\begin{equation}
\begin{split}
\mathbf{v}_{k} = \operatorname{Readout}\!\bigl(\{h_{i}\}_{i\in G_{k}},\;\{h_{i\rightarrow j}\}_{(i,j)\in G_{k}}\bigr).
\end{split}
\end{equation}

Ultimately, the model obtains the embedding set $\{\mathbf{v}_{1},\mathbf{v}_{2},\dots,\mathbf{v}_{n}\}$ as a high-level structural representation of the scene graph.  
Taking Fig.~\ref{fig3-1}(b) as an example, the model first detects the regions ``girl'', ``car'', ``tree'', and ``road''; after processing by the reasoning module, it constructs structured semantic edges such as ``girl-walking-road'', ``car-beside-girl'', ``car-behind-car'', and ``tree-under-car'', and generates attribute labels like ``black-car'' and ``silver-car''.  
According to the classification scores, several key nodes are selected and their relational sub-graphs are combined; the resulting structure-aware semantic vectors are then encoded for use by the subsequent fusion module.
\subsubsection{Textual Scene Graph Representation}
To achieve structured modeling of objects and their semantic relations in textual descriptions, we adopt a textual scene graph construction pipeline in the language modality based on syntactic parsing and GRU.This module consists of four main stages: scene graph parsing, semantic concept encoding, object–attribute graph attention, and object–object relation graph attention.

\textbf{Scene Graph Parsing.} 
Given a natural-language description $y$, we first employ a rule-based parser grounded in dependency parsing to convert the sentence into a structured scene-graph representation $G=(V,E)$.  
The graph explicitly encodes the semantic entities and their relations in the text, thereby providing a clearer and more salient topology that facilitates subsequent cross-modal semantic alignment and fusion.

More concretely, the node set $V = O \cup A$ comprises object nodes $O=\{o_i\}$ and attribute nodes $A=\{a_i\}$, whereas the edge set $E = E_{OA} \cup E_{OO}$ consists of object–attribute edges $E_{OA}\subseteq O\times A$ and object–object relation edges $E_{OO}\subseteq O\times O$.  
The parser first identifies semantic entities (noun phrases) and attribute modifiers (adjective phrases) in the input text, and then uses semantic predicates (such as prepositional or verb phrases) to construct the relations between entities.  
To improve parsing accuracy, special rules are designed for pronoun resolution, plural‐entity decomposition, and numeral normalisation, ensuring that the generated scene graph attains higher semantic fidelity and structural granularity.

\textbf{Semantic Concept Encoding.}
Each node or edge in the scene graph usually corresponds to a multi-word phrase, e.g., ``silver car" or ``walking on road".  
To effectively capture inter-word semantic dependencies and transform these structured phrases into continuous semantic vectors, we introduce a GRU encoder.  
The GRU’s gating mechanism models sequential context while mitigating long-term information loss and gradient vanishing.

Specifically, let a phrase of length $L$ be represented by the word sequence $\{w_i\}_{i=1}^{L}$, where each word embedding is $w_i\in\mathbb{R}^{e}$.  
The GRU builds the sequence representation via the following gated iterations.  
First, the update gate $z_t$ and the reset gate $r_t$, which control how much past information is retained or forgotten, are computed as: 
\begin{equation}
\begin{split}
z_t = \sigma\!\bigl(W_z w_t + U_z h_{t-1} + b_z\bigr),\\
r_t = \sigma\!\bigl(W_r w_t + U_r h_{t-1} + b_r\bigr),
\end{split}
\end{equation}
where $\sigma(\cdot)$ denotes the sigmoid activation function, $W_z, W_r \in \mathbb{R}^{d\times e}$ and $U_z, U_r \in \mathbb{R}^{d\times d}$ are learnable weight matrices, and $b_z, b_r \in \mathbb{R}^{d}$ are bias terms. Then, the GRU computes the candidate state $\tilde{h}_{t}$ at the current time step by modulating the previous hidden state $h_{t-1}$ with the reset gate:
\begin{equation}
\begin{split}
\tilde{h}_{t} = \tanh\bigl(W_{h}\,w_{t} + U_{h}\,(r_{t} \odot h_{t-1}) + b_{h}\bigr),
\end{split}
\end{equation}
where $\odot$ denotes the element-wise product, $\tanh(\cdot)$ is the hyper\-bolic tangent activation, $W_{h}\!\in\!\mathbb{R}^{d\times e}$ and $U_{h}\!\in\!\mathbb{R}^{d\times d}$ are learnable weight matrices, and $b_{h}\!\in\!\mathbb{R}^{d}$ is a bias vector. Finally, the update gate linearly interpolates the candidate state with the previous state to obtain the current hidden state: 
\begin{equation}
\begin{split}
h_{t} = (1 - z_{t}) \odot h_{t-1} + z_{t} \odot \tilde{h}_{t}.
\end{split}
\end{equation}

After the above gated iterations, the semantic representation of the phrase is given by the hidden state at the last time step: 
\begin{equation}
\begin{split}
h_{L} = \operatorname{GRU}(w_{1}, w_{2}, \dots, w_{L}),
\end{split}
\end{equation}
where $h_{L}\!\in\!\mathbb{R}^{d}$ is the final phrase embedding that preserves word-order information and contextual semantics.  
This embedding is subsequently used as the initial feature of the corresponding node or edge in the textual scene graph.

\textbf{Object–Attribute Graph Attention.}
The visual semantic representation of an object is often significantly influenced by its attribute modifiers; for instance, the word ``red’’ in ``red car’’ saliently enhances the semantics of ``car.’’  
Therefore, on the object–attribute sub-graph $G_{OA}=(O, E_{OA})$, we impose a graph attention network to realise feature aggregation between each object node $o_{i}\!\in\!O$ and its corresponding attribute nodes $a_{j}\!\in\!A$.  
Let the initial features be $h_{i}$.  
The updated node representation is computed as: 
\begin{align}
h_{i}^{\,'} &= \sum_{j\in\mathcal{N}(i)} \alpha_{ij}\,W_{g}h_{j}, \\
\alpha_{ij} &= \frac{\exp\bigl(\operatorname{LeakyReLU}\bigl(\mathbf{q}^{\top}[W_{g}h_{i}\,\Vert\,W_{g}h_{j}]\bigr)\bigr)}
                    {\sum_{k\in\mathcal{N}(i)}\exp\bigl(\operatorname{LeakyReLU}\bigl(\mathbf{q}^{\top}[W_{g}h_{i}\,\Vert\,W_{g}h_{k}]\bigr)\bigr)}, 
\end{align}
where $\mathbf{q}$ is a learnable attention vector, $W_{g}$ is a shared linear projection, $\Vert$ denotes vector concatenation, and $\mathcal{N}(i)$ is the set of attribute nodes connected to object $o_{i}$.  
This attention mechanism adaptively assigns different weights to different attributes, thereby enhancing the semantic expressiveness of the object node.  
The updated object features are denoted as $e_{i}$ and are later used as entity representations.

\textbf{Object–Object Relation Graph Attention.}
After obtaining the entity-level node features, we further consider the interactions among entities to capture richer structural semantics.  
On the object–relation subgraph $G_{OO}=(O, E_{OO})$, we introduce a context-enhanced edge mechanism, in which the original relation feature $r_{ij}$ is concatenated with the representations of object nodes $e_{i}$ and $e_{j}$ to form the context-augmented edge feature: 
\begin{equation}
\begin{split}
r_{ij}^{\,\prime} = [r_{ij}\,\Vert\,e_{i}\Vert\,e_{j}].
\end{split}
\end{equation}

To explicitly model the directionality of entity interactions, we define the active-role neighbor set $\operatorname{Act}(i)$ of node $i$ as the set of nodes connected by outgoing edges from $i$, and the passive-role neighbor set $\operatorname{Pas}(i)$ as the set of nodes connected by incoming edges to $i$.

Subsequently, we aggregate messages to the subject and object nodes in both directions, realising bidirectional context interaction:
\begin{equation}
\begin{split}
e_{i}^{\,\prime} = e_{i} + \frac{1}{|\operatorname{Act}(i)|}\sum_{j\in\operatorname{Act}(i)}W_{A}r_{ij}^{\,\prime} \\
                  + \frac{1}{|\operatorname{Pas}(i)|}\sum_{j\in\operatorname{Pas}(i)}W_{P}r_{ji}^{\,\prime},
\end{split}
\end{equation}
where $\operatorname{Act}(i)$ and $\operatorname{Pas}(i)$ denote the sets of objects in the active and passive roles with respect to node $i$, and $W_{A},W_{P}\in\mathbb{R}^{d\times d}$ are learnable transformation matrices.

To compress the locally enhanced node representations into a unified graph-level vector and achieve cross-modal alignment, we introduce a GPO.  
Given the set of enhanced object features $\{\hat{e}_{i}\}_{i=1}^{N}$, the GPO aggregates them into a fixed-length global vector $\mathbf{t}\!\in\!\mathbb{R}^{d}$, which is subsequently fed into the multimodal aggregation module.

Considering the heterogeneous contributions of different objects to textual semantics, an attention-based global pooling method is adopted to adaptively weight each node and fuse the structural information:
\begin{equation}
\begin{split}
\alpha_{i} = \frac{\exp\!\bigl(w_{p}^{\top}\tanh(W_{p}\hat{e}_{i} + b_{p})\bigr)}
                  {\sum_{j=1}^{N}\exp\!\bigl(w_{p}^{\top}\tanh(W_{p}\hat{e}_{j}^{\,\prime} + b_{p})\bigr)},
t = \sum_{i}\alpha_{i}\hat{e}_{i},
\end{split}
\end{equation}
where $\hat{e}_{i}\!\in\!\mathbb{R}^{d}$ is the enhanced feature of object~$i$, $\alpha_{i}$ is its attention score, and $w_{p},W_{p}\in\mathbb{R}^{h\times d}$, $b_{p}\in\mathbb{R}^{h}$ are learnable parameters.

The resulting graph vector $\mathbf{t}$ encapsulates the comprehensive semantic information of objects, attributes, and relations in the textual scene graph, serving as the global textual representation for subsequent multimodal alignment and image-fusion modules.

\subsection{Multimodal Scene‐Graph Hierarchical Aggregation}
\label{sec:msgha}
To fuse the structural semantic information from the visual and textual modalities more effectively, this paper proposes a multimodal scene graph hierarchical aggregation method (MSGHA).  
MSGHA mitigates the feature–misalignment problem caused by the large granularity gap and semantic drift in conventional cross-modal fusion schemes.  
By explicitly capturing the correspondences among objects, region-level cues, and global-level semantics at three different scales, MSGHA enables hierarchical, fine-grained cross-modal interaction.

Given the embedded visual features, a visual semantic reconstruction module reorganises the visual embedding $\mathbf{e}_{v}$ through self-attention into three categories of representations:  
(i) object-level features, namely the three most salient object features, denoted $\mathbf{v}_{obj}\!\in\!\mathbb{R}^{3\times d}$;  
(ii) region-level features, obtained by self-attention aggregation over all region features, denoted $\mathbf{v}_{reg}\!\in\!\mathbb{R}^{1\times d}$; and  
(iii) global-level features, derived from global average pooling, denoted $\mathbf{v}_{glob}\!\in\!\mathbb{R}^{1\times d}$.  
These features are concatenated to form the visual semantic sequence: 
\begin{equation}
\begin{split}
\mathbf{V}=[\mathbf{v}_{obj};\,\mathbf{v}_{reg};\,\mathbf{v}_{glob}]\in\mathbb{R}^{5\times d}.
\end{split}
\end{equation}

Because of the characteristics of the textual dataset, the textual modality does not require re-decomposition.  
Its object-level, region-level, and global-level semantic embeddings are concatenated directly, yielding the textual semantic sequence: 
\begin{equation}
\begin{split}
\mathbf{T}=[\mathbf{t}_{obj};\,\mathbf{t}_{reg};\,\mathbf{t}_{glob}]\in\mathbb{R}^{5\times d}.
\end{split}
\end{equation}
 
For fine-grained alignment, each corresponding semantic tier is paired.  
The visual and textual features at the same tier are concatenated and then fed into an MLP to model non-linear interactions:
\begin{equation}
\begin{split}
\mathbf{f}_{i}= \operatorname{MLP}\bigl([\mathbf{v}_{i};\mathbf{t}_{i}]\bigr),\qquad 
i\in\{{obj},\,{reg},\,{glob}\},
\end{split}
\end{equation}
and the fused feature sequence becomes: 
\begin{equation}
\begin{split}
\mathbf{F}=[\mathbf{f}_{obj};\,\mathbf{f}_{reg};\,\mathbf{f}_{glob}]\in\mathbb{R}^{5\times d}.
\end{split}
\end{equation}

To strengthen the global expressive power of the fused features and mine the correlations among different scales, a query token (CLS token) $\mathbf{q}_{cls}\in\mathbb{R}^{1\times d}$ is introduced as the query vector of a Multi-Head Self-Attention (MHSA) layer.  
The attention aggregates semantic information in the cross-modal feature sequence and outputs: 
\begin{equation}
\begin{split}
\mathbf{E}= \operatorname{MHSA}(\mathbf{q}_{cls},\,\mathbf{F},\,\mathbf{F})\in\mathbb{R}^{1\times d},
\end{split}
\end{equation}
where $\mathbf{E}$ denotes the multimodal scene-graph semantic vector after hierarchical aggregation.

\subsection{Scene Graph–Driven Image Fusion}
After obtaining the multimodal scene-graph representation $\mathbf{E}$ aggregated in the previous stage, a scene graph–driven fusion module is devised to realise fine-grained fusion of infrared and visible images under semantic guidance.  
The module adaptively regulates the contribution of each modality at the pixel level, thereby markedly enhancing the structural consistency and semantic clarity of the fused image.

Specifically, the embedded infrared features $\psi_{ir}\in\mathbb{R}^{C\times H\times W}$ and the embedded visible features $\psi_{vi}\in\mathbb{R}^{C\times H\times W}$ are first fed into two independent multilayer perceptrons (MLP) for non-linear projection, producing spatially adaptive affine-fusion parameters—weight term $\mu(\psi_{ir})\in\mathbb{R}^{C\times H\times W}$ and bias term $\lambda(\psi_{vi})\in\mathbb{R}^{C\times H\times W}$:
\begin{equation}
\begin{split}
\mu(\psi_{ir})=\operatorname{MLP}_{\mu}(\psi_{ir}), \\
\lambda(\psi_{vi})=\operatorname{MLP}_{\lambda}(\psi_{vi}).
\end{split}
\end{equation}

The module assigns the infrared modality dominant guidance in the fusion process, explicitly exploiting its salient information, while the visible modality supplies complementary detail.  
The multimodal scene-graph features serve as key constraints for structural and semantic alignment, ensuring that the fused result achieves superior consistency and sharpness.  
The final fused features are obtained via the affine transformation: 
\begin{equation}
\psi_{f}= \mu(\psi_{ir})\odot\mathbf{E}+\lambda(\psi_{vi}),
\end{equation}
where $\odot$ denotes the Hadamard product and $\psi_{f}$ represents the fused feature output.

\subsection{Loss Functions}
To guide the fusion model in preserving structural details within salient regions and integrating information in background areas of multimodal images, the loss function is composed of a reconstruction term and a local contrast term.

The reconstruction term derives from a foreground–background separation strategy. Foreground and background regions are built individually, and a mask-weighting scheme steers the network to emphasise foreground content while suppressing redundant background noise.  
The overall loss comprises two components: a region-adaptive weighted reconstruction term $L_{rec}$ and a local-contrast regularisation term $L_{{ctr}}$.

Considering that the infrared image $I_{ir}$ contains more salient targets in the foreground, whereas the visible image $I_{vi}$ provides richer background and edge information, region masks $M$ and complementary weight maps are generated by a pre-trained DenseFuse network.  
Let $w_{ir}$ and $w_{vi}$ denote the region weights for infrared and visible images, respectively.  
The region-adaptive reconstruction loss is formulated as: 
\begin{equation}
\begin{aligned}
L_{fg} &= \alpha \left\| M \odot w_{ir} \odot (\hat{I}_f - I_{ir}) \right\|_2^2 \\
              &\quad + \beta \left\| M \odot w_{vi} \odot (\hat{I}_f - I_{vi}) \right\|_2^2, \\
L_{bg} &= \gamma \left\| (1 - M) \odot (\hat{I}_f - I_{vi}) \right\|_2^2, \\
L_{rec} &= L_{fg} + L_{bg}, 
\end{aligned}
\tag{25}
\end{equation}
where $\odot$ denotes the Hadamard product, $\hat{I}_{f}$ is the fused image, and $\alpha,\beta,\gamma$ are tunable parameters, set to $2.2,\,1.2,\,1.0$ to balance the contributions of background regions.

To enhance the representation of edge intensity and texture contrast, the local-contrast term $L_{ctr}$ is introduced, reinforcing consistency between the fused image and the source images at the local structural level:
\begin{equation}
\begin{split}
L_{ctr}
= \eta \, \bigl\lVert \sigma(\hat{I}_{f})-\max\bigl(\sigma(I_{ir}),\,\sigma(I_{vi})\bigr)\bigr\rVert_{1},
\end{split}
\end{equation}
where $\sigma(\cdot)$ denotes the local standard deviation within a $9\times9$ window, and $\eta$ is a weighting coefficient set to $0.3$.

Combining the region-adaptive reconstruction loss with the local-contrast term, the final loss function is expressed as: 
\begin{equation}
\begin{split}
\mathcal{L}_{MAFL} = L_{rec} + L_{ctr}.
\end{split}
\end{equation}

\section{Experiment}
This section first describes the experimental settings, including the datasets, evaluation metrics, comparison methods, and implementation details. Subsequently, the proposed method is compared with the current state-of-the-art infrared and visible image fusion methods and scene-graph approaches. Then, an ablation study is conducted to analyse the contribution of each key component, followed by downstream tasks that validate the practicability and generalizability.

\subsection{Experimental Setting}
\textbf{Datasets.}
Extensive evaluations are carried out on three representative infrared and visible fusion datasets, namely LLVIP\,\cite{jia2021llvip}, TNO\,\cite{toet2017tno} and RoadScene\,\cite{xu2020fusiondn}, together with the classical Harvard Medical dataset\,\cite{IF-FILM}.  
Following previous work\,\cite{liu2024infrared}, the official training–testing splits are adopted. 2000 LLVIP pairs with hierarchical textual annotations are employed for model training\,\cite{TextFusion}, and the remaining 250 pairs constitute the LLVIP test set.  An additional 30 TNO pairs and 50 RoadScene pairs, all with hierarchical textual annotations, are selected as independent test sets to verify the cross-dataset generalisation capability of the proposed model.

\textbf{Evaluation metrics.}
To comprehensively evaluate the visual quality and information-preservation ability of the fused images, several widely used reference-free indices are employed: Edge Preservation Information Transfer Factor (Qabf), Structural Similarity (SSIM), Visual Information Fidelity (VIF), Average Gradient(AG), Spatial Frequency(SF), Mutual Information(MI) and Peak Signal-to-Noise Ratio(PSNR). For a holistic view, the average ranking of each method over all metrics is reported as the overall index mRank.

\textbf{Comparative Methods.}
Eight state-of-the-art multimodal image-fusion approaches are selected for comparison, including NestFusion\,\cite{nestfuse}, SwinFusion\,\cite{swinfusion}, MUFusion\,\cite{mufusion}, TextFusion\,\cite{TextFusion}, DAFusion\,\cite{DAFusion}, SpTFuse\,\cite{SpTFuse} and IF-FILM\,\cite{IF-FILM}.

\textbf{Training details.}
All methods are trained under identical conditions. Our framework is implemented in PyTorch and trained on an NVIDIA GeForce RTX 3090 GPU. During training, the learning rate of the fusion network is set to $1.0\times10^{-4}$, optimisation is performed with Adam, and the batch size is 16. All input images are uniformly cropped to $256\times256$.

\begin{figure}[tb]
	{\centering\includegraphics[width=0.5\textwidth]{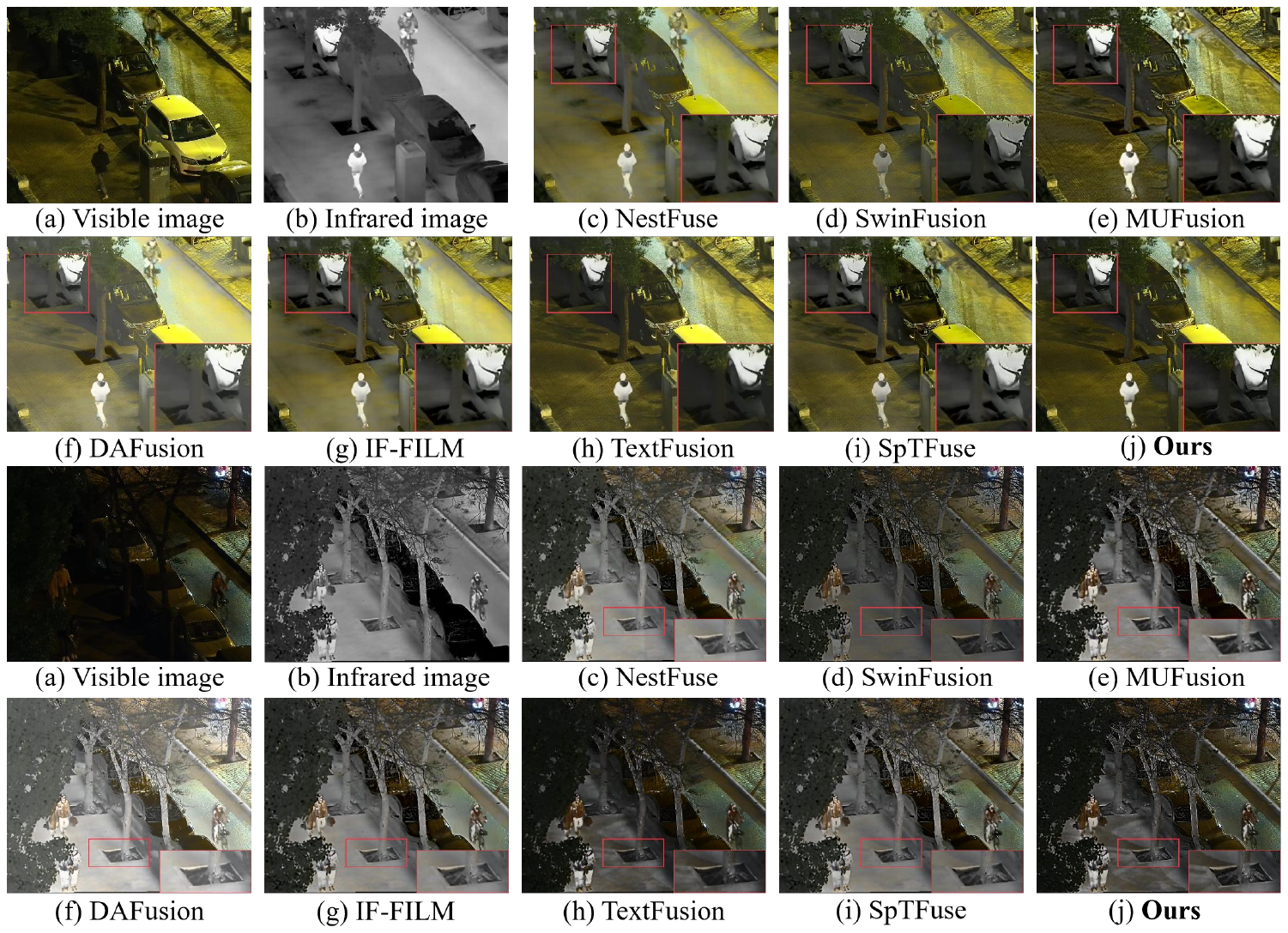}}
    \caption{Qualitative results of different methods of image fusion on LLVIP dataset images.}
	\label{fig_3}
\end{figure} 
\begin{table}[ht]
\centering
\caption{Quantitative results of different methods of image fusion on LLVIP dataset images.(Optimal: bold; 2nd-best: underlined; mRank denotes the average rank across all evaluation metrics).}
\label{tab_1}
\setlength{\tabcolsep}{3.0pt}
\renewcommand{\arraystretch}{1.05}
\begin{tabular}{lcccccccc}
\toprule
\multirow{2}{*}{Methods} & \multicolumn{8}{c}{LLVIP Dataset} \\
\cmidrule(lr){2-9}
& mRank $\downarrow$ & Qabf & SSIM & VIF & AG & SF & MI & PSNR \\
\midrule
NestFuse      & 5.714 & 0.468 & 0.562 & 0.607 & 4.684 & 12.019 & \textbf{4.005} & 18.968 \\
SwinFusion    & \underline{3.285} & \textbf{0.653} & 0.558 & 0.815 & \underline{6.888} & \underline{16.353} & \underline{3.899} & 17.816 \\
MUFusion      & 4.000 & 0.489 & 0.583 & \textbf{1.117} & 6.762 & 13.507 & 2.446 & 20.556 \\
DAFusion      & 5.143 & 0.496 & 0.489 & \underline{0.910} & 6.014 & 14.813 & 3.121 & 12.019 \\
SpTFuse       & 4.000 & 0.529 & 0.569 & 0.881 & 6.761 & 15.028 & 2.170 & \underline{20.947} \\
IF-FILM       & 7.428 & 0.235 & 0.537 & 0.673 & 4.391 & 8.566 & 3.238 & 17.239 \\
TextFusion    & 3.857 & 0.543 & \underline{0.591} & 0.683 & 6.030 & 14.966 & 2.881 & \textbf{21.406} \\
\rowcolor{gray!15}
\textbf{Ours} & \textbf{2.571} & \underline{0.620} & \textbf{0.596} & 0.803 & \textbf{7.422} & \textbf{17.869} & 2.951 & 20.105 \\
\bottomrule
\end{tabular}
\end{table}

\begin{figure}[tb]
	{\centering\includegraphics[width=0.5\textwidth]{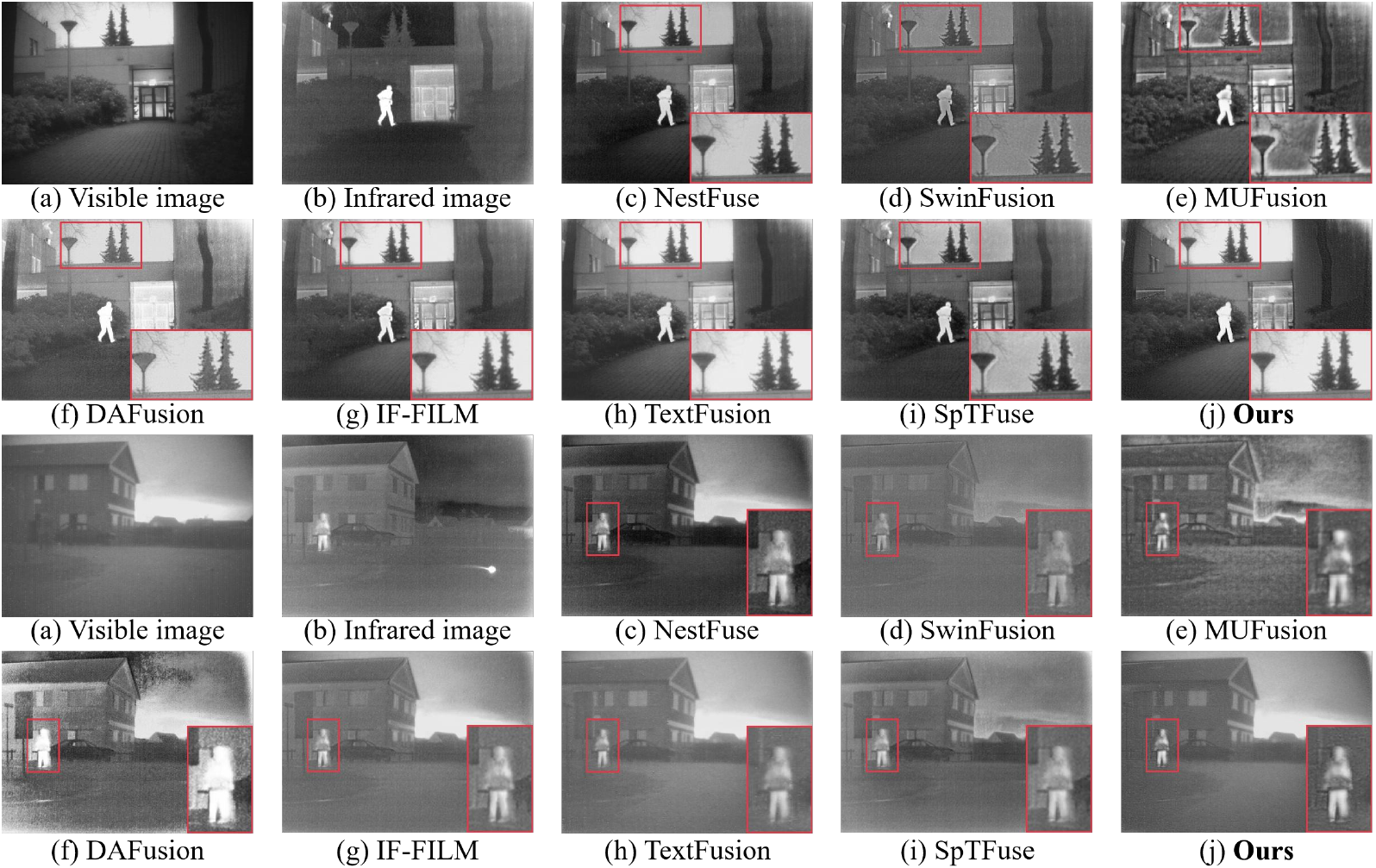}}
    \caption{Qualitative results of different methods of image fusion on TNO dataset images.}
	\label{fig_4}
\end{figure} 
\begin{table}[t]
\centering
\caption{Quantitative comparison on the \textbf{TNO} dataset.  
Optimal values are \textbf{bold}; second–best values are \underline{underlined}.  
mRank denotes the average rank (lower is better).}
\label{tab_2}
\setlength{\tabcolsep}{3.0pt}
\renewcommand{\arraystretch}{1.05}
\begin{tabular}{lcccccccc}
\toprule
\multirow{2}{*}{Method} & \multicolumn{8}{c}{TNO Dataset} \\
\cmidrule(lr){2-9}
 & mRank $\downarrow$ & Qabf & SSIM & VIF & AG & SF & MI & PSNR \\
\midrule
NestFuse      & \underline{3.429} & \textbf{0.432} & 0.473 & 0.771 & 5.234 & 10.354 & \textbf{3.261} & 19.226 \\
SwinFusion    & 4.000 & 0.421 & \underline{0.487} & 0.709 & 5.893 & 11.154 & \underline{3.246} & 16.610 \\
MUFusion      & 4.000 & 0.365 & 0.467 & \textbf{1.782} & \underline{6.759} & 10.125 & 1.945 & \underline{19.393} \\
DAFusion      & 4.000 & 0.375 & 0.454 & \underline{1.540} & \textbf{8.102}  & \textbf{14.965} & 2.702 & 15.674 \\
SpTFuse       & 4.714 & \underline{0.429} & 0.454 & 0.852 & 5.280 & 8.880 & 1.666 & \textbf{21.450} \\
IF-FILM       & 5.429 & 0.382 & 0.468 & 0.824 & 5.065 & 8.965 & 2.578 & 18.223 \\
TextFusion    & 4.571 & \textbf{0.432} & \textbf{0.520} & 0.677  & 5.243  & 9.896  & 2.916 & 17.783 \\
\rowcolor{gray!15}
\textbf{Ours} & \textbf{2.857} & \textbf{0.432} & \textbf{0.520} & 0.715  & 5.961 & \underline{11.540} & 3.077 & 18.312  \\
\bottomrule
\end{tabular}
\end{table}

\subsection{Comparisons with State-of-the-Art Methods}
\textbf{Evaluation on LLVIP.} 
Figure~\ref{fig_3} presents a qualitative comparison of fusion results produced by several advanced methods on the LLVIP testset. The visual comparison shows that different methods apply distinct modality biases in their fusion strategies. NestFuse, DAFusion and IF-FILM clearly favour the infrared modality, markedly enhancing thermal radiation in foreground regions such as pedestrians and vehicles. However, they preserve far fewer details from the visible image, causing loss of background texture and overall colour. By contrast, SwinFusion, MUFusion, SpTFuse and TextFusion focus on retaining the structure and colour information of the visible image, yet in some scenes they express hot targets less effectively, which diminishes semantic recognition under low-illumination conditions. Our method achieves a superior balance between modalities and excels at detail preservation.  Benefiting from the visual‐textual cross-modal scene-graph representation and the hierarchical multi-head attention mechanism, our method aligns semantic entities and their relations in infrared and visible images with high precision and realizes pixel-level fine-grained fusion through the affine unit. In annotated regions, its fusion of foreground thermal signals (for example, pedestrians) and background textures (such as roads and trees) surpasses all other methods, and the overall output exhibits higher contrast, sharper edges and more natural structural coherence, demonstrating the strong cross-modal semantic perception and fusion capability of our method.

According to the quantitative results in Table~\ref{tab_1}, our method attains leading performance on several key metrics.  It ranks first in Qabf and SSIM, confirming pronounced advantages in structural retention and semantic consistency.  On VIF, our method outperforms TextFusion, NestFuse and IF-FILM, further proving its ability to convey image information effectively. Our method also achieves excellent scores on AG and SF, validating its effectiveness in detail enhancement and contrast improvement.

Although NestFuse obtains the highest MI score, its limitations in structural fidelity and overall quality indicate that over-reliance on infrared information sacrifices colour content.  In contrast, our method balances complementary modality information, reconstructs structural and textural details more faithfully, and matches TextFusion on PSNR while clearly surpassing NestFuse.  Overall, our method records the best average ranking mRank among all methods, demonstrating unified superiority in global robustness and local detail preservation under bright conditions.

\textbf{Evaluation on TNO.}
Figure~\ref{fig_4} presents the fusion results produced by various methods on the TNO dataset, which contains multiple pairs of infrared and visible images taken in complex outdoor environments and therefore provides a stringent test of degradation robustness.  
As shown in Figure~\ref{fig_4}, NestFuse \cite{nestfuse}, MUFusion \cite{mufusion}, DAFusion \cite{DAFusion} and IF-FILM \cite{IF-FILM} generate results strongly biased toward the infrared modality: thermal cues of foreground targets (such as pedestrians) are significantly enhanced, yet background or edge textures from the visible image are blurred. 
SwinFusion \cite{swinfusion}, SpTFuse \cite{SpTFuse} and TextFusion \cite{TextFusion} behave more evenly in detail preservation and achieve a certain balance between the chromatic information of the two modalities. However, they still suffer from insufficient contrast and edge blurring in some foreground regions. By contrast, our method yields more favourable fusion effects in every scene.  In the first scene type shown in Figure~\ref{fig_4}, our method clearly highlights the thermal signatures of salient pedestrians while simultaneously retaining the contour and texture details of buildings and trees; in the second scene type, the luminance gradation of our method is richer and colour transitions are natural, faithfully restoring the spatial structure of the original scene. These improvements originate from the cross-modal scene-graph framework introduced by our model, which captures the latent relations between visual objects and textual entities and thus enhances the model’s capacity to generalise.

According to the quantitative results in Table~\ref{tab_2}, our method obtains the highest scores on the key metrics Qabf and SSIM, demonstrating outstanding performance in information preservation and structural fidelity. For SF, our method also exceeds IF-FILM, TextFusion and SpTFuse by a notable margin, further evidencing its superiority in edge sharpness and detail richness.  The overall metric mRank ranks our method first among all compared methods, indicating that the model achieves the most stable and colour-balanced fusion across multiple indices.  Capable of maintaining complete information and structural consistency under unknown degradation, our method exhibits the best potential for real-world applications.
\begin{figure}[tb]
	{\centering\includegraphics[width=0.5\textwidth]{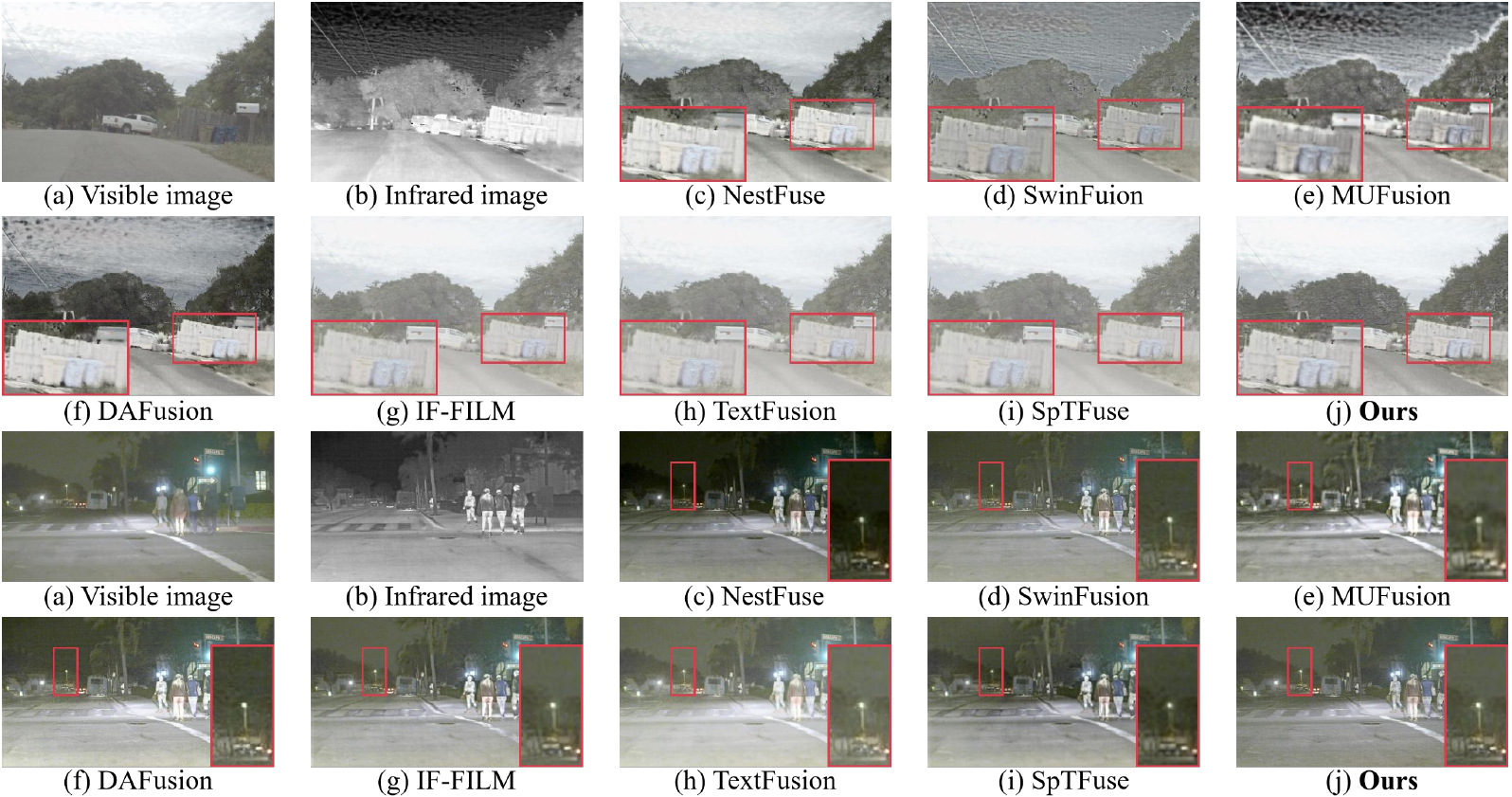}}
    \caption{Qualitative results of different methods of image fusion on RoadScene dataset images.}
	\label{fig_5}
\end{figure} 

\begin{table}[t]
\centering
\caption{Quantitative comparison on the \textbf{RoadScene} dataset.  
Optimal values are \textbf{bold}; second–best values are \underline{underlined}.  
mRank denotes the average rank (lower is better).}
\label{tab_3}
\setlength{\tabcolsep}{2.6pt}
\renewcommand{\arraystretch}{1.05}
\begin{tabular}{lcccccccc}
\toprule
\multirow{2}{*}{Method} &
\multicolumn{8}{c}{RoadScene Dataset} \\
\cmidrule(lr){2-9}
 & mRank $\downarrow$ & Qabf & SSIM & VIF & AG & SF & MI & PSNR \\
\midrule
NestFuse      & \underline{4.286} & \underline{0.493} & 0.558             & 0.718             & 6.399            & 12.630           & 3.484           & 15.244            \\
SwinFusion    & 5.286              & 0.466              & \underline{0.562} & 0.625             & 6.057            & 12.125           & \underline{3.540} & 13.570           \\
MUFusion      & 4.714              & 0.355              & 0.397             & \textbf{1.219}    & 8.011            & 12.751           & 2.467           & \underline{19.082} \\
DAFusion      & \textbf{3.286}     & 0.468              & 0.534             & 0.862 & \textbf{8.860}   & \underline{15.197} & 3.457           & 17.471            \\
SpTFuse       & 5.286              & 0.454              & 0.491             & 0.777             & 7.214            & 12.075           & 2.068           & \textbf{20.043}   \\
IF-FILM       & \textbf{3.286}     & \textbf{0.499}     & 0.527             & \underline{0.906} & \underline{8.359} & \textbf{15.264}  & 3.262           & 15.330            \\
TextFusion    & 6.571              & 0.438              & 0.550             & 0.467             & 5.140            & 10.446           & 3.460           & 14.661            \\
\rowcolor{gray!15}
\textbf{Ours} & \textbf{3.286}     & 0.482              & \textbf{0.565}    & 0.638             & 6.545            & 13.037           & \textbf{3.922}  & 15.903            \\
\bottomrule
\end{tabular}
\end{table}
\textbf{Evaluation on RoadScene.}
Figure~\ref{fig_5} shows the representative fusion results produced by several advanced methods on the RoadScene dataset.  This dataset mainly contains daytime traffic scenes under strong illumination, featuring high brightness, complex reflections and numerous occlusions. As illustrated in Figure~\ref{fig_5}, the results of SwinFusion\,\cite{swinfusion}, MUFusion\,\cite{mufusion} and DAFusion\,\cite{DAFusion} render foreground vehicles and pedestrian thermal targets rather clearly, but the backgrounds are often over–saturated, and contrast, saturation and colour balance are excessively enhanced, causing perceptual bias, information loss and noise. IF-FILM\,\cite{IF-FILM} and SpTFuse\,\cite{SpTFuse} deliver relatively balanced brightness yet still provide insufficient detail for edge structures in the backgrounds.  TextFusion\,\cite{TextFusion} preserves more RGB colour and edge information of the visible image; however, its expression of infrared targets is inadequate, and some hot targets cannot be effectively highlighted.
Compared with the competing methods, our method delivers consistently superior fusion quality across diverse target regions and background areas. In the first‐row example of Figure~\ref{fig_5}, our method not only accurately enhances the thermal signatures of foreground vehicles and pedestrians but also preserves key structural details such as building outlines and illumination–reflection boundaries. At the same time, the method successfully avoids the halo diffusion commonly observed in competing approaches; the fused image exhibits natural colours, sharp edges and well-articulated details, fully demonstrating its ability to generalise across distribution shifts. This performance is attributable to the cross modal scene-graph architecture that we design, which maintains structural modelling of scene entities and multimodal relation reasoning even under complex high-illumination scenarios, thereby producing robust fusion features.

Table~\ref{tab_3} summarises the quantitative evaluation on the RoadScene dataset. Because this dataset is never used for training, the reported results reveal each method’s ability to generalise across temporal spans and environmental conditions. The data show that our method yields the strongest overall performance. In terms of SSIM, our method surpasses SwinFusion, NestFuse and TextFusion, demonstrating a clear advantage in structural preservation and modality alignment. It also attains the highest MI score, markedly outperforming TextFusion and DAFusion, which indicates that our method integrates complementary information from the source images more effectively. For both Qabf and SF, our method achieves competitive values, confirming its capacity to maintain edge sharpness and gradient detail. Taken together, our method ranks first in the average metric mRank, reflecting its balanced superiority in global robustness and local detail retention under bright-scene conditions.

\begin{figure}[tb]
	{\centering\includegraphics[width=0.5\textwidth]{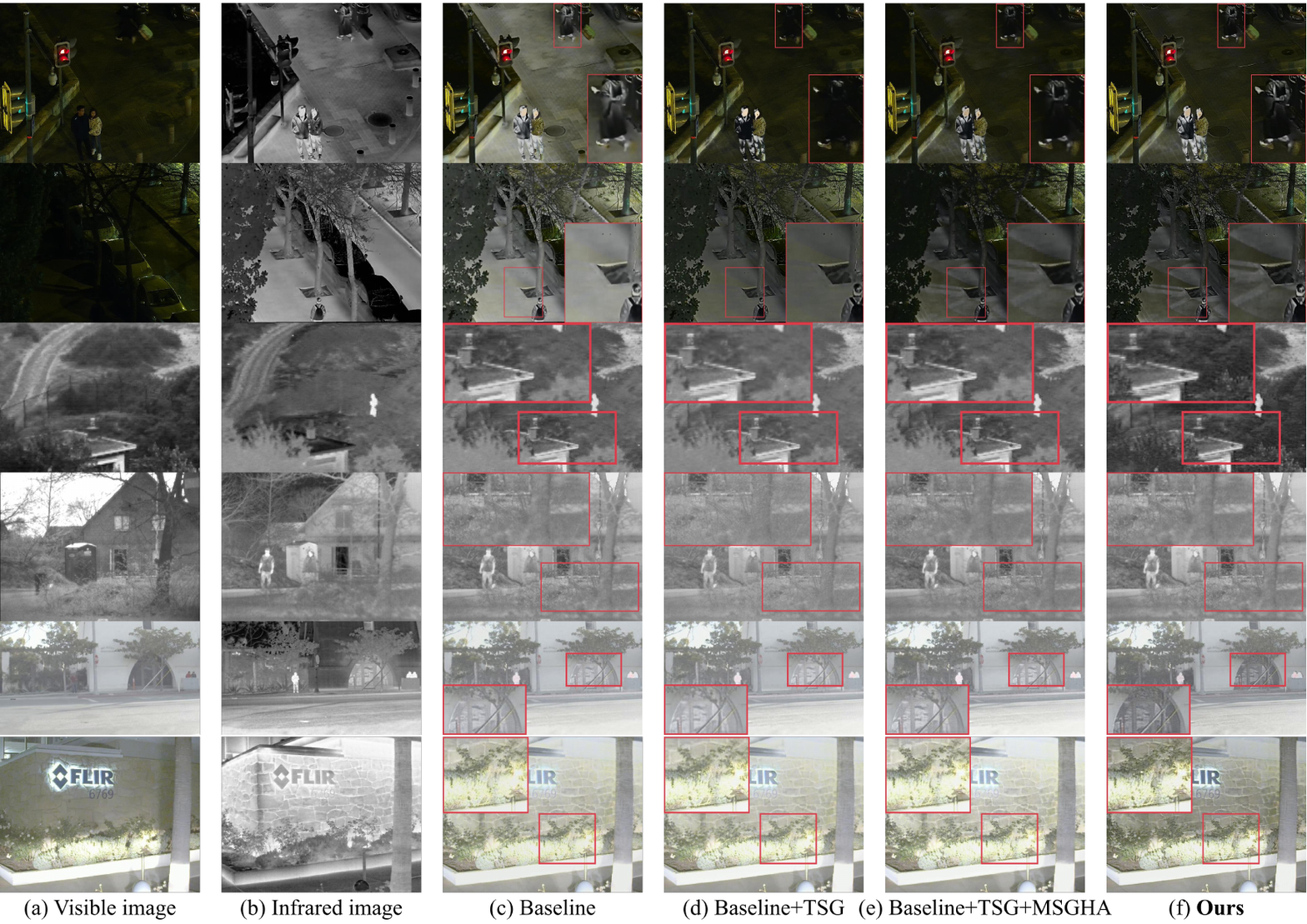}}
    \caption{Qualitative results of ablation experiments on the LLVIP, TNO and RoadScene datasets.}
	\label{fig_6}
\end{figure} 

\renewcommand{\arraystretch}{1.3}
\begin{table}[!t]
  \centering
  \caption{Quantitative results of ablation experiments on the LLVIP dataset. (Optimal: \textbf{bold}; 2nd-best: \underline{underlined}; mRank denotes the average rank across all evaluation metrics).}
  \label{tab_4}
  \scriptsize
    \setlength{\tabcolsep}{1pt}
  \begin{tabular}{cccccccccccc}
    \toprule
    mRank$\downarrow$ & Qabf$\uparrow$  & SSIM$\uparrow$ & VIF$\uparrow$ & AG$\uparrow$ & SF$\uparrow$ & MI$\uparrow$ & PSNR$\uparrow$ & Baseline & TSG & MSGHA & VSG \\
    \midrule
    3.143 & 0.452 & \underline{0.538} & 0.655 & 5.047 & 14.083 & \textbf{3.040} & 17.191 & \checkmark & & & \\
    3.000     & 0.464 & 0.527 & \underline{0.718} & \underline{5.471} & \underline{14.607} & 2.930 & 16.222 & \checkmark & \checkmark & & \\
    \underline{2.571} & \underline{0.471} & 0.535 & 0.676 & 5.177 & 14.461 & \underline{3.020} & \underline{17.193} & \checkmark & \checkmark & \checkmark & \\
    \textbf{1.286} & \textbf{0.620} & \textbf{0.596} & \textbf{0.803} & \textbf{7.422} & \textbf{17.869} & 2.951 & \textbf{20.105} & \checkmark & \checkmark & \checkmark & \checkmark \\
    \bottomrule
  \end{tabular}
\end{table}
\begin{table}[!t]
  \centering
  \caption{Quantitative results of ablation experiments on the TNO dataset. (Optimal: \textbf{bold}; 2nd-best: \underline{underlined}; mRank denotes the average rank across all evaluation metrics).}
  \label{tab_5}
  \scriptsize
    \setlength{\tabcolsep}{1pt}
  \begin{tabular}{cccccccccccc}
    \toprule
    mRank$\downarrow$ & Qabf$\uparrow$ & SSIM$\uparrow$ & VIF$\uparrow$ & AG$\uparrow$ & SF$\uparrow$ & MI$\uparrow$ & PSNR$\uparrow$ & Baseline & TSG & MSGHA & VSG \\
    \midrule
    2.714 & 0.384                     & 0.462                & 0.594    & 4.538    & 8.738               & \underline{3.405} & \underline{17.183} & \checkmark &      &      &      \\
    3.571 & 0.372                     & \underline{0.464}    & 0.526                & 4.267                & 8.476               & 3.060            & 16.645            & \checkmark & \checkmark &      &      \\
    \underline{2.143} & \underline{0.401}            & \underline{0.464}    & \underline{0.620}    & \underline{4.566}    & \underline{9.219}   & \textbf{3.575}   & 14.728            & \checkmark & \checkmark & \checkmark &      \\
    \textbf{1.286}    & \textbf{0.432}              & \textbf{0.520}       & \textbf{0.715}       & \textbf{5.961}       & \textbf{11.540}     & 3.077            & \textbf{18.312}   & \checkmark & \checkmark & \checkmark & \checkmark \\
    \bottomrule
  \end{tabular}
\end{table}
\begin{table}[!t]
  \centering
  \caption{Quantitative results of ablation experiments on the RoadScene dataset. (Optimal: \textbf{bold}; 2nd-best: \underline{underlined}; mRank denotes the average rank across all evaluation metrics).}
  \label{tab_6}
  \scriptsize
  \setlength{\tabcolsep}{1pt}
  \begin{tabular}{cccccccccccc}
    \toprule
    mRank$\downarrow$ & Qabf$\uparrow$ & SSIM$\uparrow$  & VIF$\uparrow$  & AG$\uparrow$  & SF$\uparrow$  & MI$\uparrow$  & PSNR$\uparrow$  & Baseline & TSG & MSGHA & VSG \\
    \midrule
    3.714 & 0.423 & 0.543 & 0.483 & 5.200 & 10.276 & 3.866 & \underline{15.364} & \checkmark & & & \\
    3.000     & 0.438 & 0.546 & 0.501 & 5.208 & 10.391 & \underline{3.955} & 14.723 & \checkmark & \checkmark & & \\
    \underline{2.000}     & \underline{0.473} & \underline{0.549} & \underline{0.579} & \underline{5.741} & \underline{11.826} & \textbf{4.112} & 14.728 & \checkmark & \checkmark & \checkmark & \\
    \textbf{1.286} & \textbf{0.482} & \textbf{0.565} & \textbf{0.638} & \textbf{6.545} & \textbf{13.037} & 3.922 & \textbf{15.903} & \checkmark & \checkmark & \checkmark & \checkmark \\
    \bottomrule
  \end{tabular}
\end{table}
\renewcommand{\arraystretch}{1.05}  

\subsection{Ablation Study}
In the ablation study, a series of experiments with progressively added structural components is designed to investigate how the multimodal scene graph and the loss function influence the final performance.  Qualitative and quantitative results on the LLVIP, TNO and RoadScene datasets are presented in Figure~\ref{fig_6} and Tables~\ref{tab_4}, \ref{tab_5} and \ref{tab_6}.

\textbf{Textual Scene Graph.}  On the baseline model, a textual scene graph is introduced to explore the effect of structured conceptual semantics from the text modality on fusion quality.  As shown by the second rows of Table~\ref{tab_4} and Table~\ref{tab_6}, adding the textual scene‐graph module brings notable gains in Qabf, VIF, AG and SF, indicating that the semantic–structural information in the textual scene graph positively contributes to visual fidelity and salient‐region modelling.  On the LLVIP and RoadScene datasets, the overall metric mRank increases by 4.55\% and 19.23\% over the baseline, confirming the effectiveness of the textual scene‐graph module in enhancing fusion quality.  However, the improvement on the structure‐sensitive index SSIM remains limited, suggesting that guidance from a single text modality is still insufficient for structural fidelity and that a more refined fusion mechanism is required.

\textbf{Multimodal Scene Graph Hierarchical Aggregation.}
To further strengthen information exchange between modalities, a hierarchical guidance mechanism is introduced.  
As shown in the third rows of Tables~\ref{tab_4}, \ref{tab_5} and \ref{tab_6}, the mechanism brings a significant improvement in the perceptual index VIF compared with the text-only scene-graph model, indicating that the added module greatly enhances the visual fidelity, contrast and texture of the fused images, making them more consistent with human visual perception.  
Moreover, the AG and SF scores on the TNO and RoadScene datasets show an upward trend, demonstrating that object-level, region-level, and global-level guidance enables more precise cross-modal semantic alignment and effectively improves structural stability.
However, VIF and PSNR decrease slightly after the module is added, implying that the introduction of the visible-feature branch may cause interference in ideal luminance, edge-detail and structural consistency, and that image-quality information must be considered jointly with structural guidance in subsequent fusion.

\textbf{Visual Scene Graph.  }
A visual scene graph module is then incorporated to capture spatial and semantic relations in the visual modality; together with the textual scene graph and hierarchical guidance, it jointly optimises the fusion process.  
As shown in the last rows of Tables~\ref{tab_4}, \ref{tab_5} and \ref{tab_6}, all metrics improve markedly.  
On the LLVIP dataset, Qabf, SSIM and VIF increase by 37.17\,\%, 10.78\,\% and 22.60\,\% over the baseline, respectively, while AG and SF achieve the best overall values, with AG rising by 47.06\,\%.  
These results confirm that the combined structural interaction of textual and visual scene graphs can effectively enhance the semantic representation and structural completeness of fused images.  
In addition, the combined scheme attains the lowest mRank on the LLVIP, TNO and RoadScene datasets, fully demonstrating the comprehensive superiority of the proposed multimodal scene-graph aggregation in structural stability and fusion quality.

Through the above analysis, the necessity and effectiveness of each designed module in improving fusion quality, structural stability and semantic completeness are verified, further illustrating the rationality and practical potential of the proposed model.

\begin{figure}[tb]
	{\centering\includegraphics[width=0.5\textwidth]{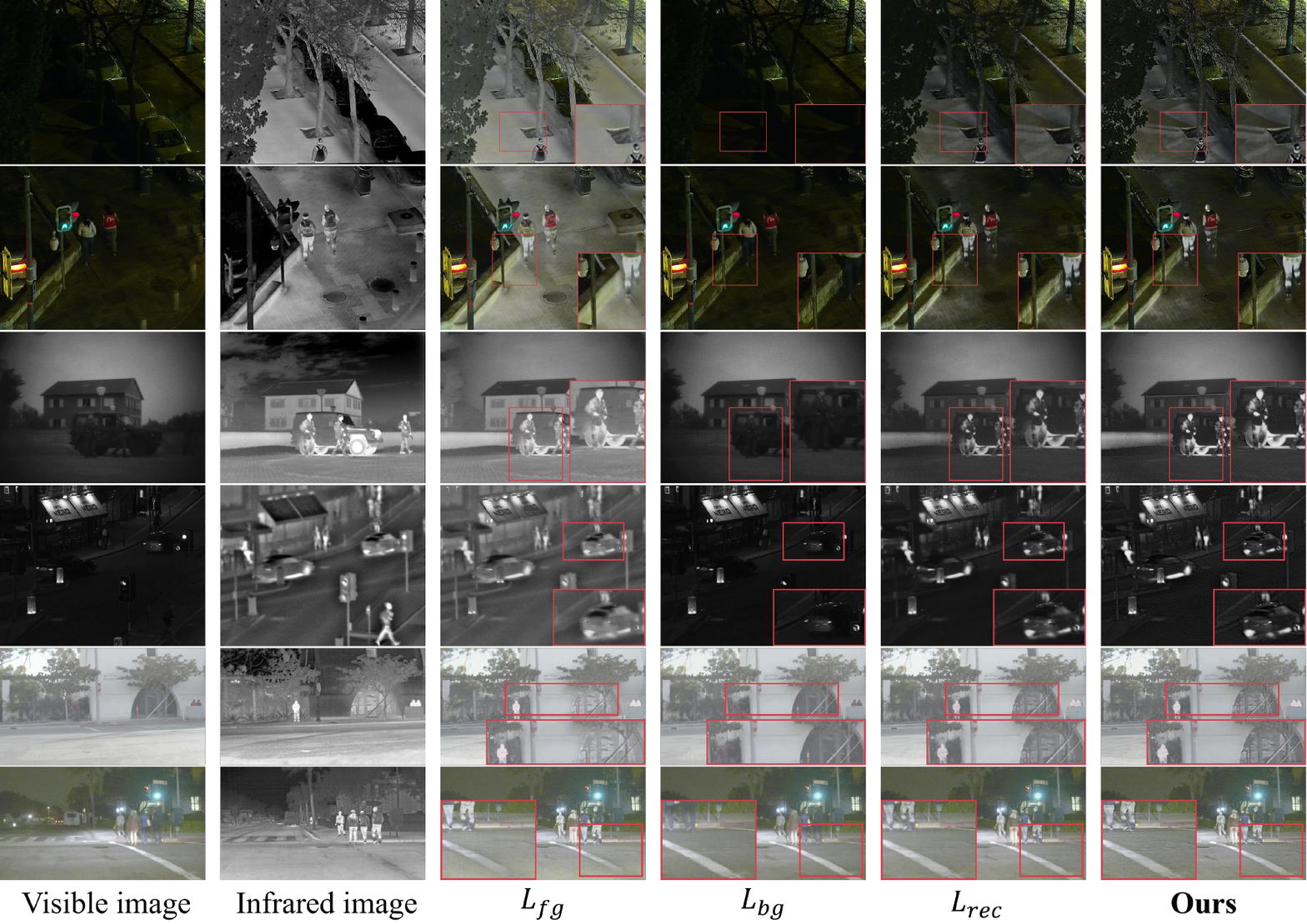}}
    \caption{Qualitative results of loss function ablation experiments on the LLVIP, TNO and RoadScene datasets.}
	\label{fig_7}
\end{figure} 
\begin{table}[!t]
  \centering
  \caption{Quantitative results of loss function ablation experiments on the LLVIP dataset. (Optimal: \textbf{bold}; 2nd-best: \underline{underlined}; mRank denotes the average rank across all evaluation metrics).}
  \label{tab_7}
  \scriptsize
  \setlength{\tabcolsep}{2.0pt}
  \begin{tabular}{ccccccccccc}
    \toprule
    mRank$\downarrow$ & Qabf$\uparrow$ & SSIM$\uparrow$ & VIF$\uparrow$ & AG$\uparrow$ & SF$\uparrow$ & MI$\uparrow$ & PSNR$\uparrow$ & $L_{{fg}}$ & $L_{{bg}}$ & $L_{{ctr}}$ \\
    \midrule
    3.571 & 0.425 & 0.531 & 0.529 & 4.417 & 12.402 & \underline{4.718} & 17.043 & \checkmark & & \\
    \underline{2.286} & \textbf{0.625} & 0.552 & 0.573 & \underline{5.918} & \underline{15.804} & \textbf{5.128} & 16.902 & & \checkmark & \\
    2.429 & 0.554 & \underline{0.594} & \underline{0.631} & 5.879 & 15.043 & 3.124 & \textbf{20.545} & \checkmark & \checkmark & \\
    \textbf{1.714} & \underline{0.620} & \textbf{0.596} & \textbf{0.803} & \textbf{7.422} & \textbf{17.869} & 2.951 & \underline{20.105} & \checkmark & \checkmark & \checkmark \\
    \bottomrule
  \end{tabular}
\end{table}
\begin{table}[!t]
  \centering
  \caption{Quantitative results of loss function ablation experiments on the TNO dataset. (Optimal: \textbf{bold}; 2nd-best: \underline{underlined}; mRank denotes the average rank across all evaluation metrics).}
  \label{tab_8}
  \scriptsize 
  \setlength{\tabcolsep}{2.0pt} 
  \begin{tabular}{ccccccccccc}
    \toprule
    mRank$\downarrow$ & Qabf$\uparrow$ & SSIM$\uparrow$ & VIF$\uparrow$ & AG$\uparrow$ & SF$\uparrow$ & MI$\uparrow$ & PSNR$\uparrow$ & $L_{{fg}}$ & $L_{bg}$ & $L_{ctr}$ \\
    \midrule
    3.571 & 0.368 & 0.454 & 0.503 & 4.008 & 8.017 & \underline{3.404} & 17.505 & \checkmark & & \\
    2.571 & \textbf{0.458} & 0.490 & 0.577 & 4.802 & 9.536 & \textbf{4.190} & 16.514 & & \checkmark & \\
    \underline{2.143} & 0.429 & \textbf{0.521} & \underline{0.624} & \underline{4.912} & \underline{9.585} & 2.203 & \textbf{18.734} & \checkmark & \checkmark & \\
    \textbf{1.714} & \underline{0.432} & \underline{0.520} & \textbf{0.715} & \textbf{5.961} & \textbf{11.540} & 3.077 & \underline{18.312} & \checkmark & \checkmark & \checkmark \\
    \bottomrule
  \end{tabular}
\end{table}
\begin{table}[!t]
  \centering
  \caption{Quantitative results of loss function ablation experiments on the RoadScene dataset. (Optimal: \textbf{bold}; 2nd-best: \underline{underlined}; mRank denotes the average rank across all evaluation metrics).}
  \label{tab_9}
  \scriptsize
  \setlength{\tabcolsep}{2.0pt}
  \begin{tabular}{ccccccccccc}
    \toprule
    mRank$\downarrow$ & Qabf$\uparrow$ & SSIM$\uparrow$ & VIF$\uparrow$ & AG$\uparrow$ & SF$\uparrow$ & MI$\uparrow$ & PSNR$\uparrow$ & $L_{{fg}}$ & $L_{bg}$ & $L_{ctr}$ \\
    \midrule
    3.143 & 0.442 & 0.556 & 0.536 & \underline{5.384} & \underline{10.803} & 3.720 & 15.034 & \checkmark & & \\
    3.000     & \textbf{0.536} & 0.514 & 0.520 & 4.949 & 10.299 & \textbf{6.010} & 15.043 & & \checkmark & \\
    \underline{2.286} & 0.463 & \textbf{0.566} & \underline{0.558} & 5.235 & 10.574 & \underline{3.939} & \underline{15.435} & \checkmark & \checkmark & \\
    \textbf{1.571} & \underline{0.482} & \underline{0.565} & \textbf{0.638} & \textbf{6.545} & \textbf{13.037} & 3.922 & \textbf{15.903} & \checkmark & \checkmark & \checkmark \\
    \bottomrule
  \end{tabular}
\end{table}

\textbf{Loss Function.}
To verify the contribution of the local-contrast regularisation term $L_{{ctr}}$ to fusion performance, an ablation study is conducted on the LLVIP, TNO and RoadScene datasets.  Figure~\ref{fig_7} presents representative qualitative comparisons.  When only the basic reconstruction loss ($L_{{fg}}$ or $L_{{bg}}$) is applied, the fused images exhibit blurred fine details and indistinct edges in both foreground and background regions.  After $L_{{ctr}}$ is introduced, the edge sharpness of targets, structural detail representation and visual contrast increase markedly; as highlighted by the red boxes, pedestrians and vehicle wheels become clearer, background details grow richer and noise is effectively suppressed.

Quantitative results in Tables~\ref{tab_7}, \ref{tab_8} and \ref{tab_9} further demonstrate the efficacy of $L_{{ctr}}$.  On the LLVIP dataset, the configuration with $L_{{ctr}}$ attains the best mRank and surpasses other settings on the key metrics Qabf, VIF and AG, improving over the baseline reconstruction loss by 45.88\,\%, 51.80\,\% and 68.03\,\%, respectively.  In cross-dataset tests on TNO and RoadScene, the $L_{{ctr}}$ configuration still maintains superior performance; on the TNO dataset, mRank reaches 1.714, and VIF and SF rise by 42.15\,\% and 43.94\,\%.  These results show that $L_{{ctr}}$ not only enhances local contrast and structural depiction in fused images but also significantly improves the model’s robustness and generalisation under various degradation scenarios.

\begin{figure*}[htb]
	{\centering\includegraphics[width=1\textwidth]{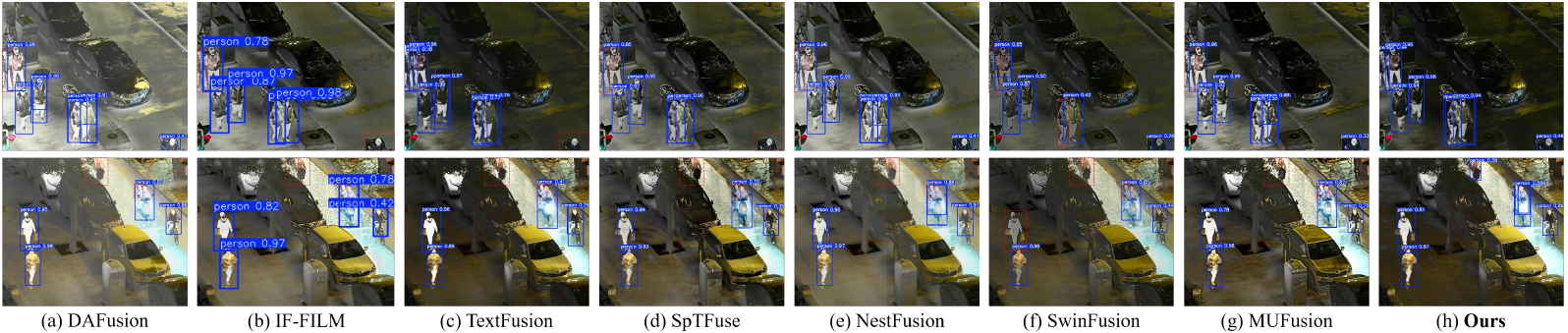}}
    \caption{Qualitative results of loss function ablation experiments on the LLVIP, TNO and RoadScene datasets.
    }
	\label{fig_8}
\end{figure*} 
\begin{table}[!t]
  \centering
  \caption{Quantitative results of pedestrian detection on the LLVIP dataset. (Optimal: \textbf{bold}; 2nd-best: \underline{underlined}).}
  \label{tab_10}
  \scriptsize
  \setlength{\tabcolsep}{9.6pt}
  \renewcommand{\arraystretch}{1.05}
  \begin{tabular}{lcccc}
    \toprule
    Method & AP@0.5 & AP@0.7 & AP@0.9 & mAP@[0.5:0.95] \\
    \midrule
    NestFuse    & 0.845 & \underline{0.748} & 0.123 & 0.529 \\
    SwinFusion  & 0.833 & 0.685 & 0.137 & 0.527 \\
    MUFusion    & 0.734 & 0.603 & 0.109 & 0.467 \\
    DAFusion    & \underline{0.853} & \underline{0.748} & 0.166 & 0.531 \\
    IF-FILM     & 0.736 & 0.653 & \textbf{0.189} & 0.492 \\
    SpTFuse     & 0.842 & 0.745 & \textbf{0.189} & \underline{0.537} \\
    TextFusion  & 0.836 & 0.716 & 0.154 & 0.508 \\
    \rowcolor{gray!15}
    \textbf{Ours}       & \textbf{0.880} & \textbf{0.751} & \underline{0.181} & \textbf{0.558} \\
    \bottomrule
  \end{tabular}
\end{table}

\subsection{Application}
To evaluate the usefulness of the fused images in downstream high-level vision tasks, an object-detection experiment and an image-segmentation experiment are carried out on the LLVIP dataset.  YOLOv11\,\cite{khanam2024yolov11} is adopted as the detector and is fine-tuned on the infrared and visible training sets of LLVIP; Segment-Anything\,\cite{kirillov2023segment} is used as the segmentation model.

\textbf{Pedestrian Detection. }
To verify the practicability of our method in downstream detection, the fused images are fed into the detector to identify pedestrian targets.  Figure~\ref{fig_8} displays the qualitative detection results; blue boxes denote correctly detected targets, whereas red boxes mark missed targets.  IF-FILM, TextFusion and SpTFuse exhibit missed detections in occluded or relatively low-contrast regions, such as pedestrians partially hidden by obstacles in the second row.  DAFusion, TextFusion and MUFusion perform better in bright regions (for example, reflective clothing under streetlights) but still miss distant or weakly illuminated targets.  By contrast, our method yields complete detections; it successfully identifies targets and background edges in complex scenes, showing stronger boundary perception and robustness.

Table\,\ref{tab_10} reports quantitative results for object detection, evaluated with four metrics: $\mathrm{AP}@0.5$, $\mathrm{AP}@0.7$, $\mathrm{AP}@0.9$, and $\operatorname{mAP}[0.5{:}0.95]$. Our method achieves the best performance on three of the four indicators; in particular, its $\operatorname{mAP}[0.5{:}0.95]$ score exceeds that of the visually comparable TextFusion by approximately $9.8\,\%$. It is also noteworthy that our method attains a superior $\mathrm{AP}@0.9$ score relative to every baseline except IF-FILM and SpTFuse, which indicates higher accuracy in boundary detail preservation and high confidence target localisation. We attribute these gains to the cross modal scene graph semantic fusion mechanism of our method, which effectively integrates thermal saliency from the infrared channel with structural and edge cues from the visible channel, thereby providing the detector with clearer and semantically consistent fused feature representations.

\begin{figure}[!t]
	{\centering\includegraphics[width=0.5\textwidth]{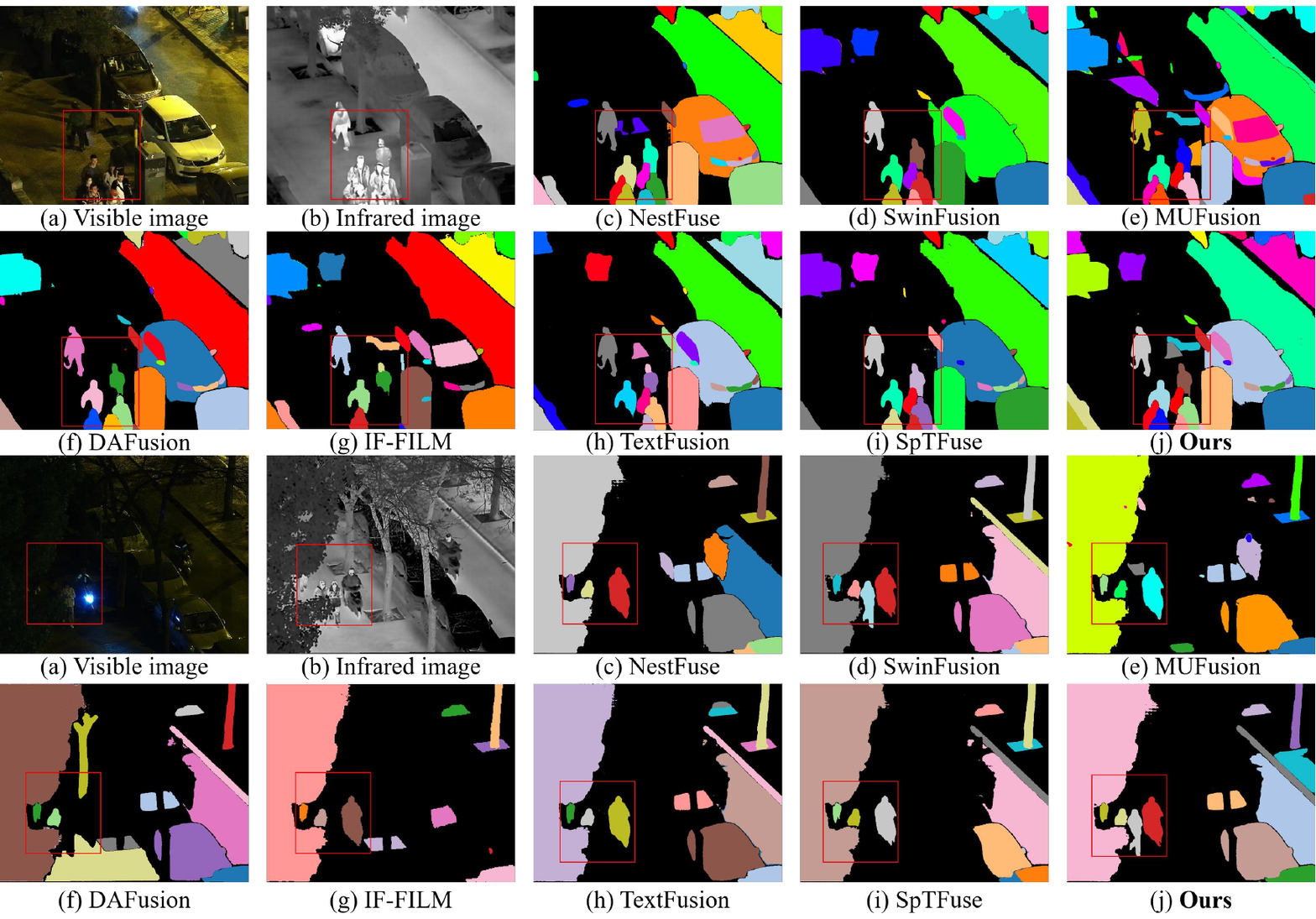}}
    \caption{Qualitative results of different methods of image segmentation on LLVIP dataset.}
	\label{fig_9}
\end{figure} 
\begin{figure}[!t]
	{\centering\includegraphics[width=0.5\textwidth]{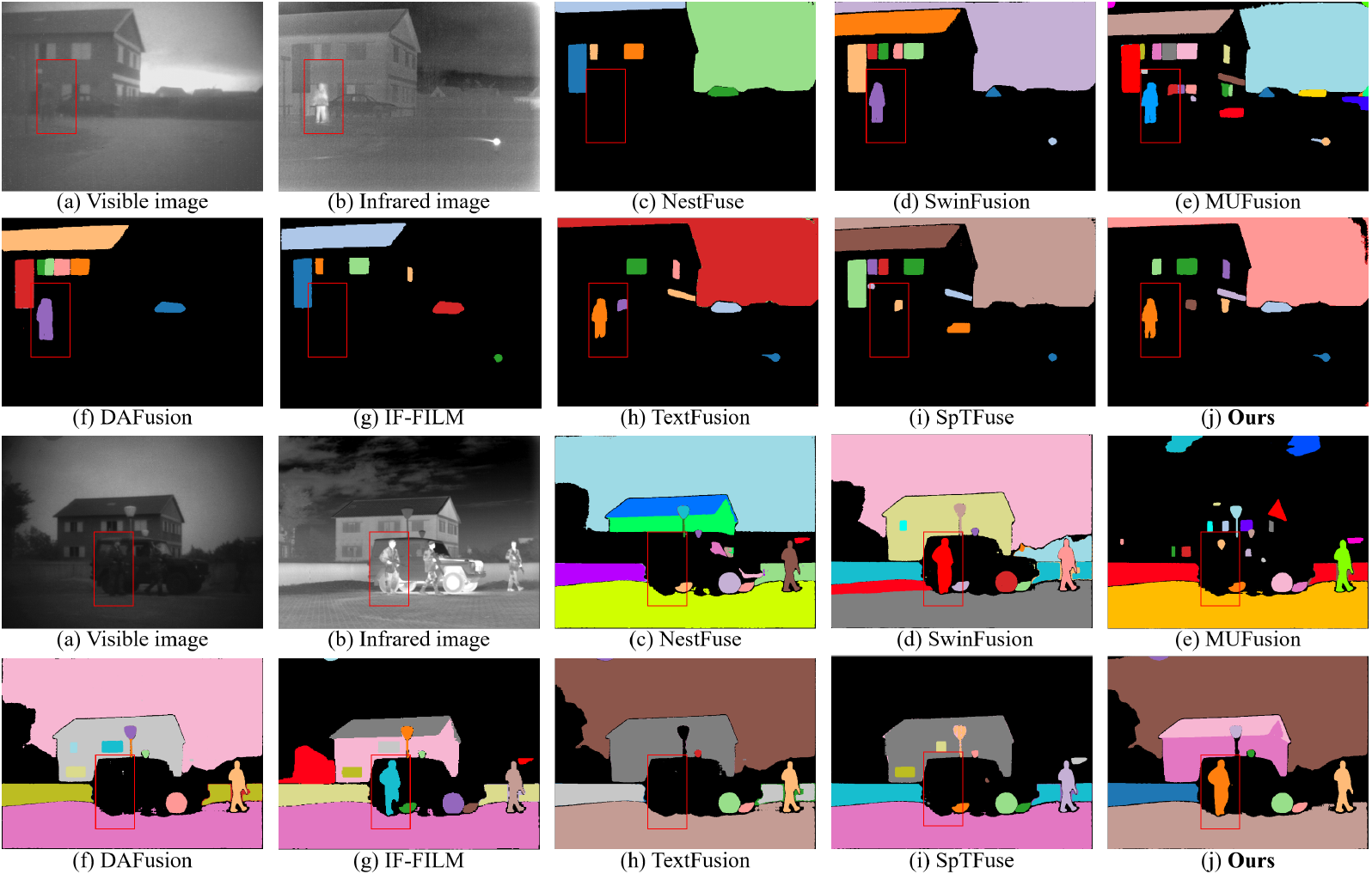}}
    \caption{Qualitative results of different methods of image segmentation on TNO dataset.}
	\label{fig_10}
\end{figure} 
\begin{figure}[!t]
	{\centering\includegraphics[width=0.5\textwidth]{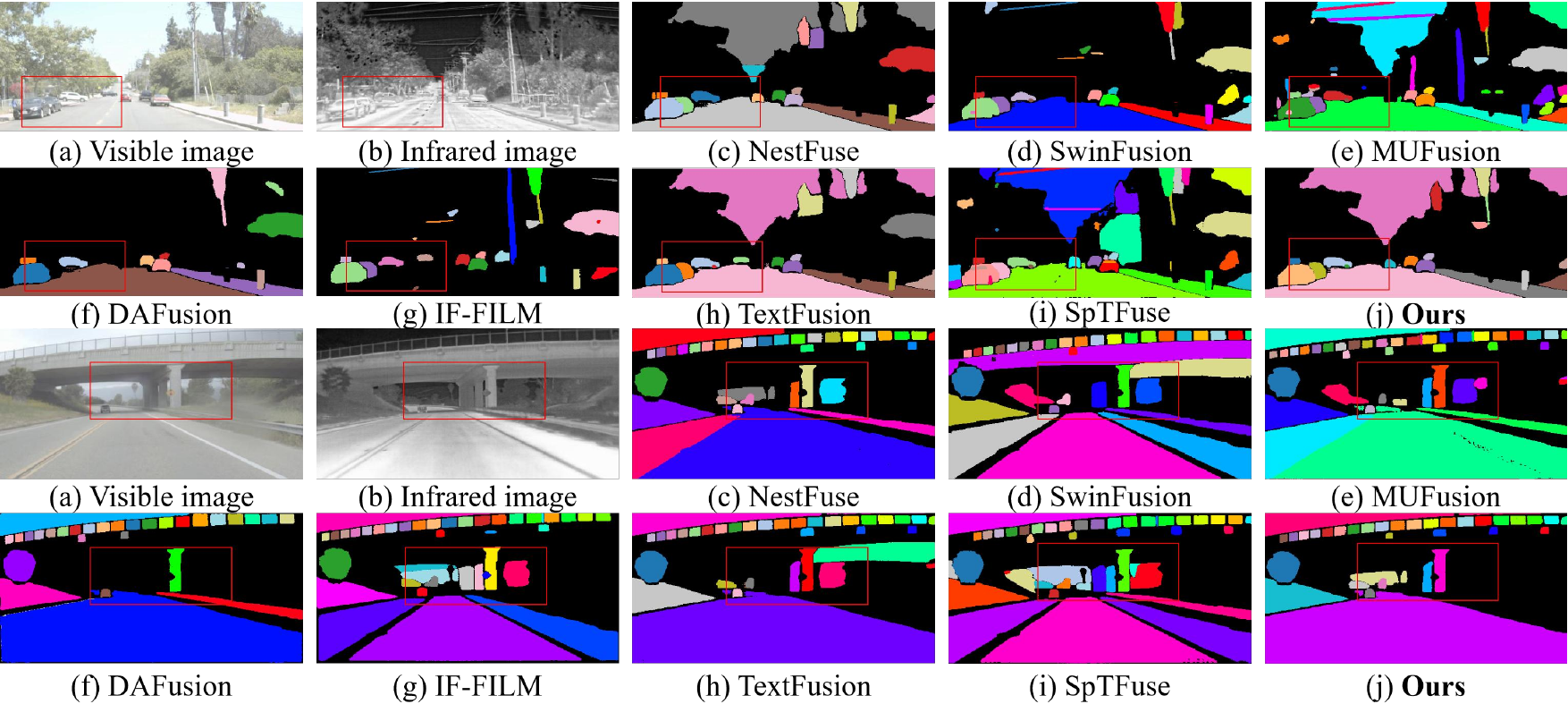}}
    \caption{Qualitative results of different methods of image segmentation on RoadScene dataset.}
	\label{fig_11}
\end{figure}

\textbf{Image Segmentation. }
To verify the adaptability of the fused images to global perception tasks, the fused outputs of each method are further fed into a semantic‐segmentation model and evaluated cross-domain on the LLVIP, TNO and RoadScene datasets.  Figures~\ref{fig_9}–\ref{fig_11} illustrate segmentation results in representative scenes, revealing marked differences in quality among the methods.

On the LLVIP dataset, SwinFusion and IF-FILM exhibit strong target-region recognition under nighttime conditions, yet their masks suffer from structural breakage and contour adhesion at edges. NestFuse and MUFusion preserve background textures well, but the masks of high-level semantic entities such as pedestrians and vehicles remain incomplete.  In contrast, our method achieves the best performance in the red-boxed areas, accurately distinguishing pedestrians from road backgrounds; its boundaries are sharp and complete, with no missed segments.

On the TNO and RoadScene datasets, our method continues to demonstrate superior structural discrimination, effectively separating complex classes such as sky, buildings and vehicles.  In occlusion regions and strong-light interference areas, its mask boundaries remain stable without noticeable shift or blur.  These results indicate that the fused images produced by our method possess higher semantic consistency and contextual separability, thereby enhancing the cross-domain adaptability of the segmentation model and improving downstream generalisation.
\begin{figure}[tb]
	{\centering\includegraphics[width=0.5\textwidth]{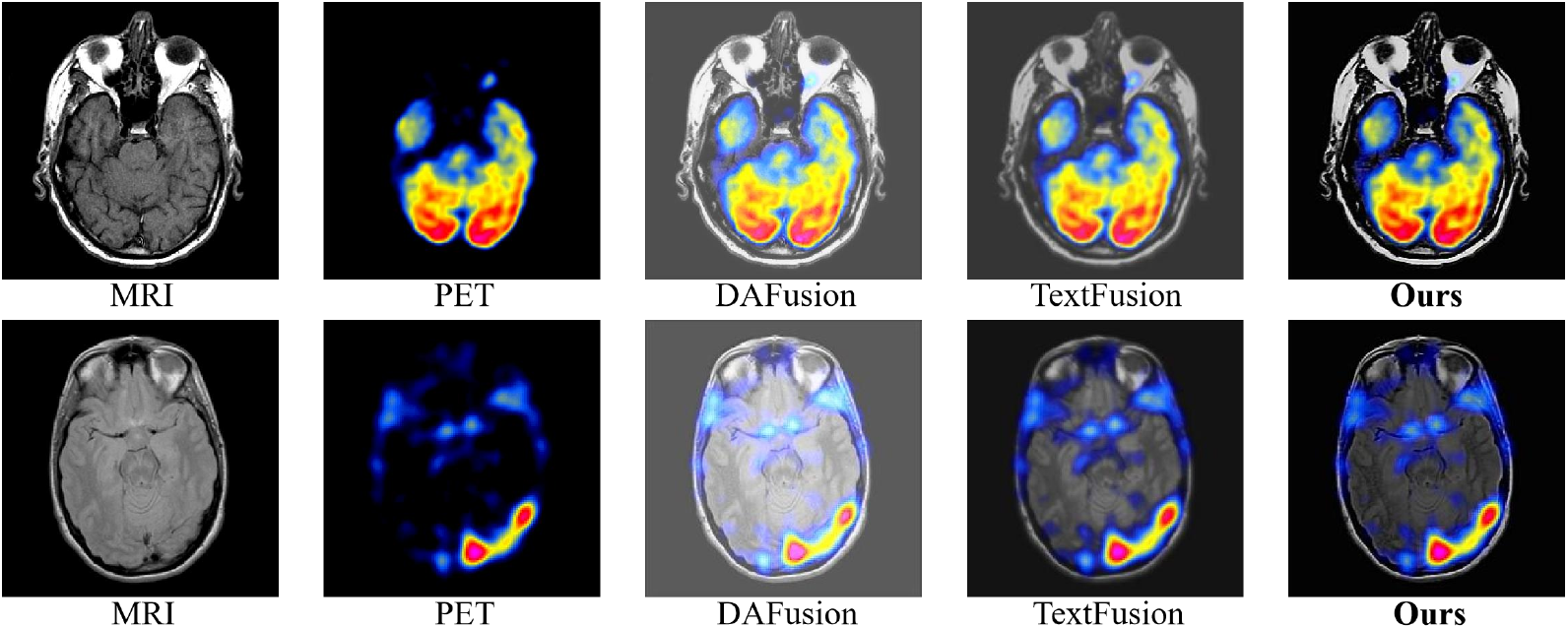}}
    \caption{Qualitative results of image fusion on the Harvard Medical dataset.}
	\label{fig_12}
\end{figure} 

\begin{table}[ht]
  \centering
  \caption{Quantitative results of image fusion on the Harvard Medical dataset. (Optimal: \textbf{bold}; 2nd-best: \underline{underlined}; mRank denotes the average rank across all evaluation metrics).}
  \label{tab_11}
  \scriptsize
  \setlength{\tabcolsep}{7.8pt}
  \renewcommand{\arraystretch}{1.05}
  \begin{tabular}{lcccccc}
    \toprule
    Method & mRank$\downarrow$ & Qabf$\uparrow$  & SSIM$\uparrow$ & VIF$\uparrow$ & SF$\uparrow$ & MI$\uparrow$ \\
    \midrule
    TextFusion & \underline{2.200} & 0.274 & \underline{0.296} & \underline{0.532} & 10.548 & \textbf{1.598} \\
    DAFusion   & 2.600 & \underline{0.484} & 0.283 & 0.514 & \underline{16.296} & 1.571 \\
    \rowcolor{gray!15}
    \textbf{Ours}      & \textbf{1.200} & \textbf{0.515} & \textbf{0.384} & \textbf{0.537} & \textbf{21.014} & \underline{1.573} \\
    \bottomrule
  \end{tabular}
\end{table}
\subsection{Generalizability}
To validate the generalizability of the proposed model, a medical-image fusion task is conducted on the Harvard Medical dataset.  Figure~\ref{fig_12} compares representative qualitative MRI–PET fusion results.  Relative to DAFusion and TextFusion, the proposed method preserves high contrast while revealing clearer organisational structures with sharper boundaries; for example, the cerebral cortex and metabolically active regions exhibit higher brightness and uniform distribution, edge details are sharper and artefacts are suppressed, thus providing improved diagnostic readability and structural completeness.

Quantitative results are summarised in Table~\ref{tab_11}.  The proposed model attains the highest Qabf and SF scores, markedly surpassing TextFusion and DAFusion, which indicates superior contrast handling and texture representation.  The model also ranks first in the overall metric mRank, demonstrating stronger capability to maintain source-image structural fidelity while enhancing key salient regions, thereby exhibiting considerable cross-domain adaptability and practical potential.

\section{Conclusion}
In this paper, we propose MSGFusion, a multimodal scene graph–guided framework for infrared and visible image fusion. It is the first to deeply integrate textual conceptual semantics with visual attributes and spatial relationships, thereby balancing high‐level semantics and low‐level details. By constructing cross‐modal scene graphs and employing a hierarchical aggregation mechanism, MSGFusion precisely aligns entities and their relations across infrared and visible inputs, achieving a deep fusion of semantic and visual information. Extensive experiments demonstrate that MSGFusion significantly outperforms state‐of‐the‐art methods in structural similarity, information fidelity, and detail preservation, and exhibits exceptional generalizability in medical image fusion and downstream tasks across various degradation scenarios. Future work will focus on developing a unified fusion model applicable to a broader range of tasks, supporting more diverse vision applications.

\section{References}

\bibliographystyle{IEEEtran}
\bibliography{reference.bib}




\end{document}